\documentclass[11pt]{article}

\usepackage{acl}

\usepackage{times}
\usepackage{latexsym}

\usepackage[T1]{fontenc}
\usepackage[utf8]{inputenc}

\usepackage{microtype}

\usepackage{inconsolata}

\usepackage{graphicx}

\usepackage{hyperref}
\usepackage{url}
\usepackage{booktabs}
\usepackage{multirow}

\usepackage{amsmath}
\usepackage{amssymb}
\usepackage{mathtools}
\usepackage{amsthm}

\usepackage[disable,textsize=tiny]{todonotes}
\usepackage[most]{tcolorbox}
\usepackage{listings}
\usepackage{array}
\usepackage{tabularx}
\usepackage{algorithm}
\usepackage{algorithmic}
\usepackage[table]{xcolor}
\usepackage{colortbl, longtable}
\usepackage{wrapfig}
\usepackage{cleveref}
\usepackage{subcaption}

\definecolor{Mycolor-red}{HTML}{FCEAEA} % light red

\definecolor{pinkframe}{HTML}{D26AA8}
\definecolor{pinktitle}{HTML}{F8D7EA}
\definecolor{pinkbg}{HTML}{FFF5FB}

\definecolor{myblueframe}{HTML}{4C8DFF} % 边框：浅蓝
\definecolor{mybluetitle}{HTML}{DCEAFF} % 标题栏：很浅蓝
\definecolor{mybluebg}{HTML}{F5F9FF}    % 内容底：极浅蓝

\definecolor{grayframe}{HTML}{7A7A7A}
\definecolor{graytitle}{HTML}{EDEDED}
\definecolor{graybg}{HTML}{F7F7F7}

\newtcblisting{promptbox}[2][]{
  enhanced,
  colframe=grayframe,
  colback=graybg,
  colbacktitle=graytitle,
  coltitle=black,
  fonttitle=\bfseries,
  title={#2},
  boxrule=0.8pt,
  arc=3pt,
  left=8pt,right=8pt,top=6pt,bottom=8pt,
  listing only,
  listing options={
    basicstyle=\ttfamily\small,
    breaklines=true,
    columns=fullflexible
  },
  #1
}

\newtcolorbox{promptboxmath}[1]{
  enhanced,
  colframe=grayframe,
  colback=graybg,
  colbacktitle=graytitle,
  coltitle=black,
  fonttitle=\bfseries,
  title={#1},
  boxrule=0.8pt,
  arc=3pt,
  left=8pt,right=8pt,top=6pt,bottom=8pt
}

\newtcblisting{steerbox}[1]{
  enhanced,
  colframe=myblueframe,
  colback=mybluebg,
  colbacktitle=mybluetitle,
  coltitle=black,
  fonttitle=\bfseries,
  title=#1,
  boxrule=0.8pt,
  arc=2pt,
  left=6pt,right=6pt,top=6pt,bottom=6pt,
  listing only,
  listing options={
    basicstyle=\ttfamily\small,
    breaklines=true,
    columns=fullflexible
  }
}

\newtcolorbox{skillcard}[1]{
  enhanced,
  colframe=pinkframe,
  colback=pinkbg,
  colbacktitle=pinktitle,
  coltitle=black,
  title=#1,
  fonttitle=\bfseries,
  boxrule=0.8pt,
  arc=3pt,
  left=8pt,right=8pt,top=6pt,bottom=8pt,
}

\theoremstyle{plain}
\newtheorem{assumption}{Assumption}[section]
\newtheorem{definition}[assumption]{Definition}

\theoremstyle{remark}

\usepackage{lineno}

\definecolor{darkblue}{rgb}{0, 0, 0.5}
\hypersetup{colorlinks=true, citecolor=darkblue, linkcolor=darkblue, urlcolor=darkblue}

\title{Toward Latent Language Model Skills Steering and Optimization: An Empirical Study}

\author{
Xunyi Jiang$^{1}$, Junda Wu$^{2}$, Yuxin Xiong$^{3}$, Sheldon Yu$^{3}$, Tong Yu$^{2}$, David Arbour$^{2}$, \\
\textbf{Ritwik Sinha$^{2}$, Julian McAuley$^{3}$, Hongyi Wen$^{1,*}$}
\\
$^{1}$New York University \quad
$^{2}$Adobe Research \quad
$^{3}$UC San Diego \\
\texttt{\{xj2511,hongyi.wen\}@nyu.edu} \quad 
\texttt{\{jundaw,tyu,arbour,risinha\}@adobe.com} \\
\texttt{\{y7xiong,ziy040,jmcauley\}@ucsd.edu} \\
\vspace{2pt}
{\small $^{*}$Corresponding author.}
}

\begin{document}
\maketitle
\begin{abstract}
Skills, as a useful abstraction for the procedural capabilities of large language models (LLMs), capture how models perform structured, multi-step reasoning and program execution.
Existing approaches typically treat skills as explicit, surface-level constructs specified through prompts or programs, leaving open the question of how such procedural capabilities are represented inside the model and whether they can be manipulated as structured objects in latent space.
In this empirical study, we investigate whether procedural LLM skills can be \emph{represented} as directions in activation space and whether vector-space operations over these directions can express skill-level behaviors.
We find that procedural skills admit a vector-space representation: individual skill directions can be activated to shift model behavior; independently extracted directions can compose to form higher-level skills.
Contrastive directions yield context-conditioned algorithmic personalization and optimization trajectories over skill directions evolve non-monotonically, with intermediate states often surpassing fully optimized solutions.
These results support a representation-level view of procedural LLM skills: they admit a latent vector-space organization that allows direct manipulation through internal interventions.
\end{abstract}

\section{Introduction}

Skills provide a central abstraction for describing the procedural capabilities of large language models (LLMs). Skills usually can be described as structured, multi-step behaviors that govern how models perform reasoning, program execution, and flexible problem solving\footnote{\url{https://en.wikipedia.org/wiki/Skill}}. 
Recent work has studied how such skills emerge, compose, and can be reused across tasks~\citep{SKiC, park2025doesrlposttraininginduce, wei2026compositionalgeneralizationllmsskill, yu2025selfimprovingagent, zhang2026memskill}.
These approaches model skills as composable reasoning operators~\citep{ComposableChainsofThought}, executable modules~\citep{zheng2025skillweaver, wang2025inducingprogrammaticskills}, or structured training units~\citep{xia2026skillrl}, but primarily treat them as explicit or surface-level constructs learned during training or specified through prompts~\citep{park2025instructskillmix, yang2025automatedskilldiscoverylanguage, alzubi2026evoskill}.
In parallel, activation steering has shown that intervening on hidden activations can control LLM behavior at inference time~\citep{DBLP:conf/iclr/LeePRMDND25, DBLP:conf/nips/CaoZC00MC24, cho2025corrsteergenerationtimellmsteering, DBLP:SAKE, li2026steeringvectorfieldscontextaware}.
However, prior steering work has largely targeted stylistic or attribute-level variation such as tone, persona, refusal, and factual associations~\citep{DBLP:conf/nips/ArditiOSPPGN24, DBLP:Tonebank}, leaving open how \emph{procedural skills} are organized inside the model.

In this work, we ask a question:
\emph{To what extent are procedural LLM skills represented as directions in activation space, and can vector-space operations over these directions express skill-level behaviors?}
We approach this question by extracting skill-conditioned directions from hidden representations with multiple steering methods, each matched to the structure of the corresponding sub-study, and by studying how these directions behave under inference-time intervention and gradient-based optimization.
Compared with prior task-vector and behavioral-steering work, we focus on procedural skills with verifiable execution, where the behavioral target is the model's internal procedure rather than a stylistic surface attribute.

\paragraph{An LLM skill, in this study.}
We use \emph{LLM skill} to refer to a procedural capability that the model applies during generation.
We constructed benchmarks \textsc{SkillSet-Math} and \textsc{SkillSet-Code}.
In \textsc{SkillSet-Math}, atomic skills are specified at the operation level, for example, adding two single digits, and composite skills are explicit compositions of them.
In \textsc{SkillSet-Code}, a skill corresponds to one element of an algorithmic strategy pair, for instance, DFS vs.\ BFS.

\textbf{Main Findings.} 
Our findings are organized by two properties: \textbf{steerability} and \textbf{optimizability}. 
At inference time, we examine whether latent interventions can activate individual skills and support their composition. 
Under optimization, we study whether latent skill representations can be adapted to different preferences and evolve under objectives. The main findings are:
\begin{enumerate}
\item Skills can be activated directly through latent steering (\Cref{sec:rq1} \textcolor{blue}{\textcircled{\scriptsize A}}).
\item Latent skill composition is effective and often outperforms prompting-based composition (\Cref{sec:rq2} \textcolor{orange}{\textcircled{\scriptsize C}}).
\item Latent skills enable skill personalization, allowing different execution strategies under varying preferences (\Cref{sec:rq3} \textcolor{teal}{\textcircled{\scriptsize P}}).
\item Skill evolution is non-monotonic, with intermediate latent states often outperforming fully optimized solutions (\Cref{sec:rq4} \textcolor{purple}{\textcircled{\scriptsize E}}).
\end{enumerate}

These results suggest that procedural capabilities in LLMs are encoded as structured latent objects that can be manipulated and adapted through internal interventions. 
This provides a new perspective on model control, shifting from surface-level prompting to representation-level skill control.

\section{Related Work}
% \paragraph{Skills in Large Language Models.}
% Recent work studies how skills emerge and compose in LLMs.
% Some approaches model skills as \emph{atomic reasoning operations}.
% For example, SKiC~\citep{SKiC} and Learning Composable Chains-of-Thought~\citep{ComposableChainsofThought} represent skills as textual operators that can be composed to solve complex problems.
% Similarly, \citet{park2025doesrlposttraininginduce} analyzes skill composition induced by reinforcement learning and defines atomic skills as elementary operators that combine to perform reasoning.
% Other work treats skills as executable modules in agent systems.
% Agent Skill Induction~\citep{wang2025inducingprogrammaticskills} represents skills as Python functions verified through re-execution, while SkillWeaver~\citep{zheng2025skillweaver} enables agents to autonomously discover reusable APIs.
% More recent work organizes skills into explicit taxonomies or training pipelines, such as skill-taxonomy guided data synthesis~\citep{wei2026compositionalgeneralizationllmsskill} and instruction mixing across skill sets~\citep{park2025instructskillmix}.
% These works focus primarily on \emph{learning} or \emph{discovering} skills during training.

\paragraph{Skills in Large Language Models.}
Recent work has examined how skills emerge and compose in large language models (LLMs).
Some approaches view skills as \emph{atomic reasoning operations}.
For example, SKiC~\citep{SKiC} and Learning Composable Chains-of-Thought~\citep{ComposableChainsofThought} represent skills as composable textual operators.
Similarly, \citet{park2025doesrlposttraininginduce} study skill composition induced by reinforcement learning and define atomic skills as elementary operators for reasoning.
Other work treats skills as executable modules in agent systems.
Agent Skill Induction~\citep{wang2025inducingprogrammaticskills} represents skills as Python functions verified by re-execution, while SkillWeaver~\citep{zheng2025skillweaver} enables agents to discover reusable APIs.
More recent work organizes skills into explicit taxonomies or structured training pipelines, including skill-taxonomy guided data synthesis~\citep{wei2026compositionalgeneralizationllmsskill} and instruction mixing across skill sets~\citep{park2025instructskillmix}.
Overall, these approaches focus on \emph{learning} or \emph{discovering} skills during training, and treat skills as explicit or surface-level constructs rather than latent representations that can be directly manipulated in the model’s internal space.

\paragraph{Skill Discovery and Skill Libraries in Agents.}
Another line of work studies skill acquisition in autonomous agents.
Automated Skill Discovery~\citep{yang2025automatedskilldiscoverylanguage} enables agents to iteratively explore environments and refine reusable capabilities through interaction and feedback.
Self-improving agents with skill libraries store executable programs that can be reused across tasks~\citep{yu2025selfimprovingagent}.
Recent work further investigates automated skill evolution, where agents iteratively refine and expand their capabilities through exploration and feedback, including failure-driven refinement mechanisms~\citep{alzubi2026evoskill}.
Memory-centric and embodied settings further study how skills evolve and transfer across tasks.
For example, MemSkill~\citep{zhang2026memskill} models memory operations as learnable skills, while SkillRL~\citep{xia2026skillrl} evolves a hierarchical skill bank during reinforcement learning.
Collectively, these approaches emphasize skills as reusable and evolving units that support generalization and efficiency in agent systems.

\paragraph{Activation Steering in Language Models.}
Activation steering methods manipulate hidden representations at inference time to control model behavior without modifying model parameters.
Prior work derives steering vectors from contrastive examples or learned representations.
Representative methods include Contrastive Activation Addition~\citep{panickssery2023steering}, Representation Engineering~\citep{representation_eng}, and Conditional Activation Steering~\citep{DBLP:conf/iclr/LeePRMDND25}, which construct or condition steering directions from contrastive prompts and conceptual axes.
More recently, CorrSteer~\citep{cho2025corrsteergenerationtimellmsteering} extracts steering directions from sparse autoencoder features, while steering vector fields introduce context-dependent control directions~\citep{li2026steeringvectorfieldscontextaware}.
Activation steering has also been applied to knowledge editing and behavioral control.
SAKE~\citep{DBLP:SAKE} performs knowledge editing via activation transport, and persona-based methods control stylistic or personality traits through activation vector algebra~\citep{chen2025personavectorsmonitoringcontrolling,pai2026billysteeringlargelanguage,feng2026personadynamiccompositionalinferencetime}.
However, existing approaches primarily focus on controlling \emph{style}, \emph{persona}, or \emph{factual attributes}, rather than procedural capabilities.
Whether activation steering can directly control \emph{skills}, for example, task execution modes, remains largely unexplored.
Rather than proposing a new steering mechanism, we address this gap with an empirical study of skill-level steering across tasks and domains.
\begin{figure*}
    \centering
    \includegraphics[width=0.8\linewidth]{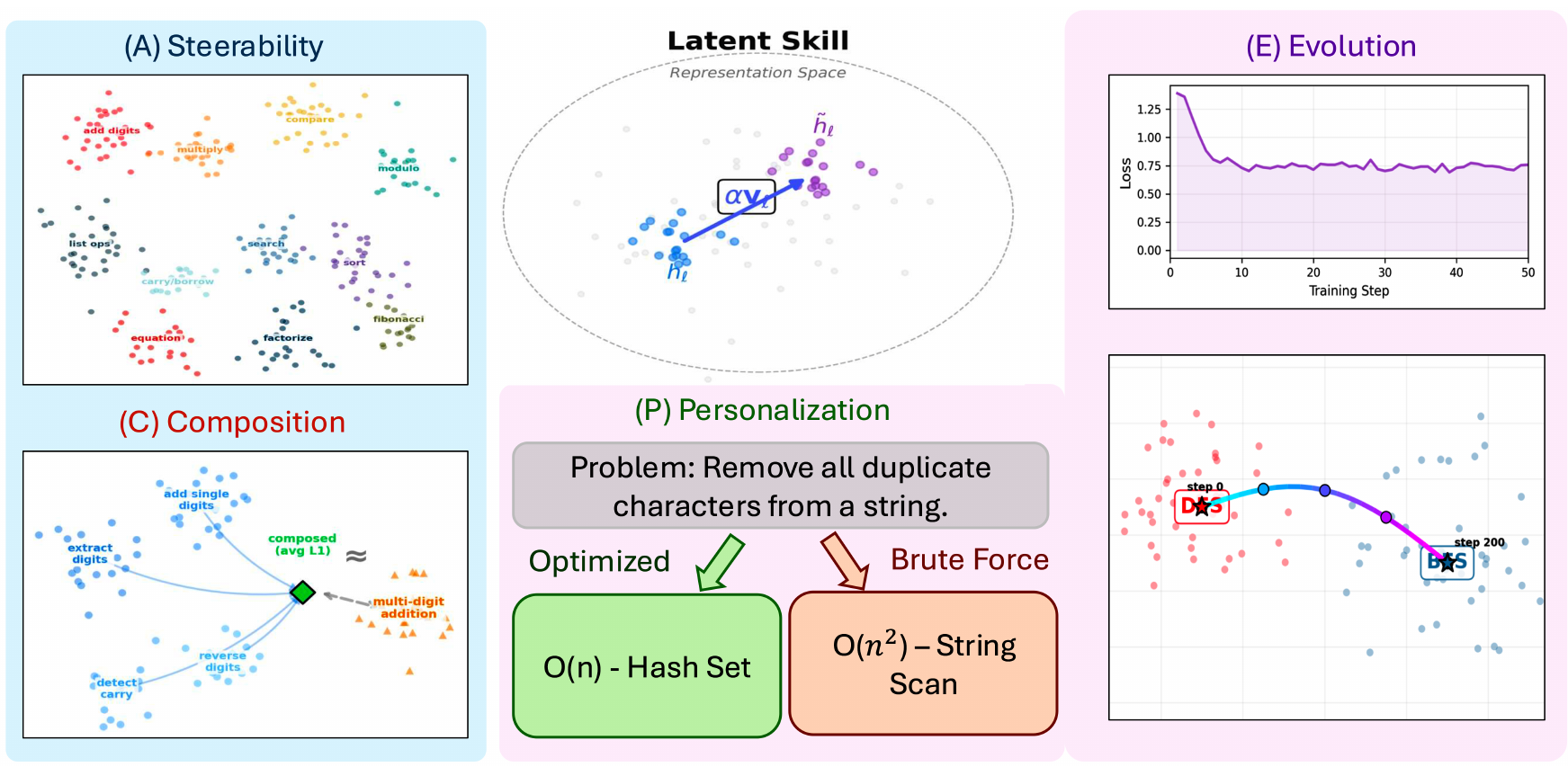}
    \caption{
    \textbf{Latent LLM Skills as Representation-Level Objects.}
    We study four observable properties of skill-conditioned directions in activation space: \textcolor{blue}{steerability}, \textcolor{orange}{composition}, \textcolor{teal}{personalization}, and \textcolor{purple}{evolution}.
    }
    \label{fig:latent_skill}
\end{figure*}

\section{Problem Formulation} \label{sec:formulation}

We formalize the object of study, \textbf{latent LLM skill directions}, as vectors in activation space that capture procedural behavior in a pretrained language model.
Our central question is whether such directions, extracted from skill-conditioned hidden representations, can be \emph{activated}, \emph{composed}, \emph{personalized}, and \emph{evolved} via internal interventions.
This perspective recasts skills as representation-level controls rather than surface-level specifications.
We then state four observable properties used throughout the paper:
\textbf{steerability \textcolor{blue}{\textcircled{\scriptsize A}} (\Cref{sec:skill-steering})} and \textbf{composition} \textcolor{orange}{\textcircled{\scriptsize C}} at inference time,
and \textbf{personalization} \textcolor{teal}{\textcircled{\scriptsize P}} and \textbf{evolution (\Cref{sec:skill-opt})} \textcolor{purple}{\textcircled{\scriptsize E}} under gradient-based optimization.

\subsection{Latent LLM Skill}
\label{sec:def-skill}

\begin{definition}[Latent LLM Skill Direction]
\label{def:latent-skill}
Let $f_\theta$ denote a pretrained language model with hidden representations
$\{h_\ell(x)\}_{\ell=1}^L$, where $h_\ell(x) \in \mathbb{R}^{d_\ell}$ is the hidden state at layer $\ell$ for input $x$, and let $\mathcal{S} \subseteq \{1,\dots,L\}$ denote the set of intervened layers.
A latent LLM skill direction is a collection of layerwise intervention vectors
\begin{equation}
v = \{v_\ell\}_{\ell \in \mathcal{S}}, \qquad v_\ell \in \mathbb{R}^{d_\ell},
\end{equation}
which, together with a scalar activation strength $\alpha \in \mathbb{R}$, defines intervened representations $\tilde{h}_\ell$ and the corresponding skill-conditioned model $f_\theta^{(v,\alpha)}$ by
\begin{equation}
\begin{aligned}
\tilde{h}_\ell(x;\alpha) &= \begin{cases} h_\ell(x) + \alpha v_\ell, & \ell \in \mathcal{S}, \\ h_\ell(x), & \ell \notin \mathcal{S}, \end{cases} \\
f_\theta^{(v,\alpha)}(x) &:= f_\theta\!\left(x;\{\tilde{h}_\ell(x;\alpha)\}_{\ell=1}^{L}\right)
\end{aligned}
\end{equation}
\end{definition}
Intuitively, $v$ specifies a direction in representation space, and $\alpha$ controls its activation strength.
Thus, a latent skill direction intervenes on hidden states during the forward pass without modifying the model parameters,
and we write $\hat{y} \sim f_\theta^{(v,\alpha)}(x)$ for the resulting output.
However, not every direction in hidden-state space is behaviorally meaningful.
We therefore introduce a validity criterion that requires a candidate direction to induce a nontrivial change in the expected task objective when activated.

\begin{definition}[Valid Skill Direction]
A latent LLM skill direction $v$ is said to be valid on a task distribution $\mathcal{D}$ if
\begin{equation} \label{eq:skill-validity}
\begin{aligned}
\mathcal{J}(v,\alpha) &:= \mathbb{E}_{x \sim \mathcal{D}} \bigl[ \mathcal{J}_0(f_\theta^{(v,\alpha)}(x)) \bigr], \\
\text{s.t.} \quad & \frac{\partial \mathcal{J}(v,\alpha)}{\partial \alpha}\Big|_{\alpha=0} \neq 0,
\end{aligned}
\end{equation}
where $\mathcal{J}_0$ is a sample-level task objective (e.g., the indicator of an incorrect answer).
Equivalently, an infinitesimal activation of $v$ produces a first-order change in the expected task behavior.
\end{definition}
We use $\mathcal{J}$ for the expected task objective throughout to avoid overloading $\mathcal{L}$, which we reserve for indexing layers (e.g., the layer set $\mathcal{S}$ above and the model depth $L$).

\subsection{Latent LLM Skill Steering} \label{sec:skill-steering}
We next study whether such skill directions can act as effective inference-time controls.
Latent activation steering enables a steering vector to be added to intermediate hidden states to modulate model behavior~\citep{subramani2022extracting,turner2023steering,panickssery2023steering}.
In our setting, however, the target of control is a procedural \emph{LLM skill}.
We therefore formulate latent skill steering as the problem of whether activating a valid skill direction can produce a measurable gain in task performance, and whether multiple such steerable directions can be further composed in latent space.

\textbf{\textcolor{blue}{\textcircled{\scriptsize A}} Steerability.}
Intuitively, steerability requires that $v$ act as a meaningful latent control direction: changing the activation strength $\alpha$ should not merely perturb hidden states,
but should improve the expected task objective over the unsteered model.
\begin{definition}[Latent Skill Steerability]
Let $v$ be a valid skill direction according to~\Cref{eq:skill-validity}. We say that $v$ is \emph{steerable} on $\mathcal{D}$ if
\begin{equation}
\exists \alpha \in \mathbb{R}
\quad \text{s.t.} \quad
\Delta(v,\alpha) := \mathcal{J}(0,0) - \mathcal{J}(v,\alpha) > 0.
\end{equation}
\end{definition}
Because the intervention is continuous in $\alpha$, this also captures graded inference-time control.
In our evaluation, we instantiate the task objective $\mathcal{J}$ as the error rate over the dataset.
Accordingly, $\Delta(v,\alpha)$ corresponds to the reduction in error rate induced by activating the latent skill direction.
Empirically, this is measured as accuracy improvement over the unsteered baseline.

\textbf{\textcolor{orange}{\textcircled{\scriptsize C}} Skill Composition.}
Once individual skill directions are steerable, we further study whether they can be jointly activated to express composite skills.
Compositional metric learning constructs richer representations by recombining learned primitives or sub-embeddings~\citep{atzmon2020causal,naeem2021learning,zheng2021deep,kaya2019deep}.
Motivated by this compositional view, we define vector-space skill composition through joint hidden-state intervention and characterize it by the resulting interaction gain between two steerable directions.
\begin{definition}[Vector-Space Skill Composition]
Let $v_i$ and $v_j$ be two steerable skill directions. We define their joint intervention by
\begin{equation}
\begin{aligned}
\tilde{h}_\ell(x;\alpha_i,\alpha_j) &= h_\ell(x) + \alpha_i v_{i,\ell} + \alpha_j v_{j,\ell}, \\
f_\theta^{(v_i,v_j;\alpha_i,\alpha_j)}(x) &:= f_\theta\!\left(x;\{\tilde{h}_\ell(x;\alpha_i,\alpha_j)\}_{\ell=1}^L\right),
\end{aligned}
\end{equation}
and their interaction gain by
\begin{equation}
\begin{aligned}
\Delta_i &:= \mathcal{J}(0,0) - \mathcal{J}(v_i,\alpha_i), \\
\Delta_{ij} &:= \mathcal{J}(0,0) - \mathcal{J}(v_i,v_j;\alpha_i,\alpha_j), \\
\Gamma_{ij} &:= \Delta_{ij} - \Delta_i - \Delta_j.
\end{aligned}
\end{equation}
We say that $v_i$ and $v_j$ admit \emph{vector-space composition} if
\begin{equation}
\exists \alpha_i,\alpha_j \in \mathbb{R}
\quad \text{s.t.} \quad
\Gamma(v_i,v_j;\alpha_i,\alpha_j) \neq 0.
\end{equation}
\end{definition}
Intuitively, $\Gamma$ measures whether the effect of jointly activating two skill directions is reducible to the sum of their individual effects.
In our evaluation, we do not explicitly estimate $\Gamma$.
Instead, we assess composition by testing whether composed skill directions improve performance on composite tasks beyond the unsteered baseline and, more importantly, beyond \emph{text-based composition} baselines.
This comparison is the central axis of our composition study: we ask whether vector-space composition expresses something that text-based composition cannot.

\subsection{Latent LLM Skill Optimization}\label{sec:skill-opt}

We next study whether latent LLM skill directions can be \emph{optimized} as continuous objects under downstream objectives~\citep{representation_eng}.
While latent skill steering asks whether a valid skill direction can be activated as an inference-time control variable,
latent skill optimization asks whether the same latent object can be improved through gradient-based updates in latent space.
To this end, let $\Phi$ denote a parameterization of a latent skill direction, with $v_{\Phi}$ denoting the corresponding intervention.
Given a downstream objective $\mathcal{J}_{\mathrm{opt}}(\Phi;\sigma)$ under contextual preference $\sigma$, we perform gradient-based updates
\begin{equation}
\begin{aligned}
\Phi^{(t+1)} &= \Phi^{(t)} - \eta \nabla_\Phi \mathcal{J}_{\mathrm{opt}}(\Phi^{(t)};\sigma), \\
\mathcal{T} &= \{\Phi^{(0)}, \Phi^{(1)}, \dots, \Phi^{(T)}\}.
\end{aligned}
\end{equation}
Based on this optimization view, we study \emph{personalized} skill directions and the \emph{evolution} of skill directions along training trajectories, complementing prior work on optimization of representation interventions~\citep{DBLP:conf/nips/WuAWGJMP24,cho2025corrsteergenerationtimellmsteering,DBLP:WuRepSteer25Nuerips,subramani2022extracting}.

\textbf{\textcolor{teal}{\textcircled{\scriptsize P}} Skill Personalization.}
If skill directions are optimizable, then conditioning optimization on different contextual preferences should yield distinct but still task-effective configurations.
We therefore formulate latent skill personalization as a preference-conditioned optimization problem in latent space, instantiated in this paper for paired algorithmic strategies (e.g., DFS vs.\ BFS).
\begin{definition}[Latent Skill Personalization]
Given an initial latent skill parameterization $\Phi^{(0)}$ and a contextual preference $\sigma$, the latent skill personalization problem is to solve
\begin{equation}
\begin{aligned}
\Phi_\sigma^* &= \arg\min_{\Phi}\, \mathcal{J}_{\mathrm{opt}}(\Phi;\sigma) \\
&\text{s.t.} \quad \exists\, \alpha \in \mathbb{R} \ \text{with} \ \Delta(v_\Phi, \alpha) > 0.
\end{aligned}
\end{equation}
For two preferences $\sigma_i \neq \sigma_j$, personalization is realized when the corresponding solutions $\Phi_{\sigma_i}^*$ and $\Phi_{\sigma_j}^*$ induce distinguishable preference-aligned behaviors while remaining task-effective.
\end{definition}
This formulation asks whether the same underlying latent skill direction can be specialized into multiple preference-conditioned variants while preserving its utility on the task.

\textbf{\textcolor{purple}{\textcircled{\scriptsize E}} Skill Evolution.}
A natural follow-up question is whether the \emph{optimization trajectory} itself contains additional structure---specifically, whether intermediate checkpoints correspond to meaningful and progressively improved skill configurations.
Prior work on representation-intervention optimization has reported non-monotonic dynamics: steering vectors can over-fit late in training, with intermediate checkpoints generalizing better than the final ones~\citep{DBLP:WuRepSteer25Nuerips,subramani2022extracting}.
The following definition is the \emph{operational} definition of skill-evolution behavior used in this empirical study; we do not claim it as a theoretical contribution, but use it to make the trajectory-level evaluation in \Cref{sec:rq4} precise.
\begin{definition}[Latent Skill Evolution]
\label{def:evolution}
Given an initial latent skill parameterization $\Phi^{(0)}$, a downstream objective $\mathcal{J}_{\mathrm{opt}}$, and the induced optimization trajectory
\begin{equation}
\begin{aligned}
\Phi^{(t+1)} &= \Phi^{(t)} - \eta \nabla_\Phi \mathcal{J}_{\mathrm{opt}}(\Phi^{(t)};\sigma), \\
\mathcal{T} &= \{\Phi^{(0)}, \Phi^{(1)}, \dots, \Phi^{(T)}\},
\end{aligned}
\end{equation}
we say the trajectory exhibits \emph{(non-monotonic) skill evolution} if there exist optimization states $\Phi^{(t_1)}$ and $\Phi^{(t_2)}$ with $t_1 < t_2$ such that
\begin{equation}
\mathcal{M}(\Phi^{(t_2)}) < \mathcal{M}(\Phi^{(t_1)}),
\end{equation}
where $\mathcal{M}$ is a downstream evaluation metric.
\end{definition}

Taken together, this formulation views a latent LLM skill direction as a representation-level object endowed with \textbf{steerability} and \textbf{optimizability}, and four observable signatures: \textcolor{blue}{\textcircled{\scriptsize A}} steerability, \textcolor{orange}{\textcircled{\scriptsize C}} composition, \textcolor{teal}{\textcircled{\scriptsize P}} personalization, and \textcolor{purple}{\textcircled{\scriptsize E}} evolution.
These four signatures define the scope of our evaluation in \Cref{sec:experiments}.

\section{Empirical Study of Vector-Space Skill Representations}
\label{sec:experiments}

Our experiments empirically probe the four observable properties of latent LLM skill directions defined in~\Cref{sec:formulation}:
\textcolor{blue}{\textcircled{\scriptsize A}} steerability,
\textcolor{orange}{\textcircled{\scriptsize C}} composition,
\textcolor{teal}{\textcircled{\scriptsize P}} personalization, and
\textcolor{purple}{\textcircled{\scriptsize E}} evolution.
We organize them into two experimental settings.

\textbf{Controlled foundational study.}
Skill directions are extracted and evaluated on our constructed benchmarks \textsc{SkillSet-Math} (RQ1, RQ2) and \textsc{SkillSet-Code} (RQ3, RQ4), where each skill is procedurally defined and the desired behavior is verifiable.

\textbf{Real-benchmark evaluation.}
The math-skill directions extracted in \textsc{SkillSet-Math} are applied to the standard math reasoning benchmarks GSM8K and MATH500, across Qwen3 models from 0.6B to 14B~\citep{yang2025qwen3}. In this real-benchmark setting, we ask whether the same vector-space directions can transfer beyond the controlled benchmark.

\subsection{Datasets and Models}

\paragraph{Controlled Skill Datasets.}
We construct two domain-specific datasets with explicit skill annotations: \textsc{SkillSet-Math} ($\mathcal{D}_{\text{math}}$) and \textsc{SkillSet-Code} ($\mathcal{D}_{\text{code}}$), with details in~\Cref{app:skillset}.

\textsc{SkillSet-Math} is a math-reasoning dataset with two levels of granularity: atomic skills (e.g., \texttt{add\_two\_single\_digits}, \texttt{detect\_carry}) and composite skills (e.g., \texttt{multi\_digit\_addition}) formed by composing multiple atomic skills.
This structure enables controlled evaluation of \textcolor{blue}{\textcircled{\scriptsize A}} steerability and \textcolor{orange}{\textcircled{\scriptsize C}} composition, since we can directly test whether single skill directions induce the intended atomic behaviors and whether independently extracted directions compose to solve higher-level tasks.

\textsc{SkillSet-Code} consists of 9 contrastive paired skills, where each pair represents two valid but distinct algorithmic approaches to the same problem (e.g., DFS vs.\ BFS).
This dataset is used to evaluate \textcolor{teal}{\textcircled{\scriptsize P}} personalization and \textcolor{purple}{\textcircled{\scriptsize E}} evolution.

Both benchmarks are controlled by design.
Isolable skills, known compositional structure, and verifiable outputs are what make the four properties measurable at all, and open-ended benchmarks do not expose this structure.
We therefore do not claim they exhaust the complexity of procedural skills in practice, but rather treat them as a testbed for establishing whether such skills have a manipulable representation-level structure.

\paragraph{Real Benchmarks.}
For the real-benchmark evaluation, the skill directions extracted on \textsc{SkillSet-Math} are transferred to GSM8K and MATH500.

\paragraph{Models.}
In the controlled foundational study, we use Qwen3-0.6B and Qwen3-1.7B.
In the real-benchmark evaluation, we use Qwen3 models ranging from 0.6B to 14B~\citep{yang2025qwen3}.

\paragraph{Skill-Direction Extraction.}
All experiments use the extraction procedures formalized in~\Cref{sec:method-extraction}: single-class PCA for math skills (RQ1 and RQ2) and contrastive mean-difference for paired algorithmic strategies (RQ3); RQ4 optimizes a parameterized skill direction along training trajectories.

\subsection{\textcolor{blue}{\textcircled{\scriptsize A}} Do skill directions applied at inference time improve LLM math reasoning?}
\label{sec:rq1}

\paragraph{Setup}
We extract one PCA-based skill direction $v$ per skill on Qwen3-0.6B, using the procedure in~\Cref{sec:method-extraction}.
Each direction is extracted at every layer and applied to the final 50\% of layers with $\alpha = 2.0$.
Evaluation is on the test split of $\mathcal{D}_{\text{math}}$ (100 samples per skill), compared to the zero-shot baseline $\mathrm{Acc}(0,0;\mathcal{D}_{\text{math}})$.

\textbf{Result~1: Skill directions shift behavior on \textsc{SkillSet-Math} for Qwen3-0.6B.}
\begin{figure*}
    \centering
    \includegraphics[width=\linewidth]{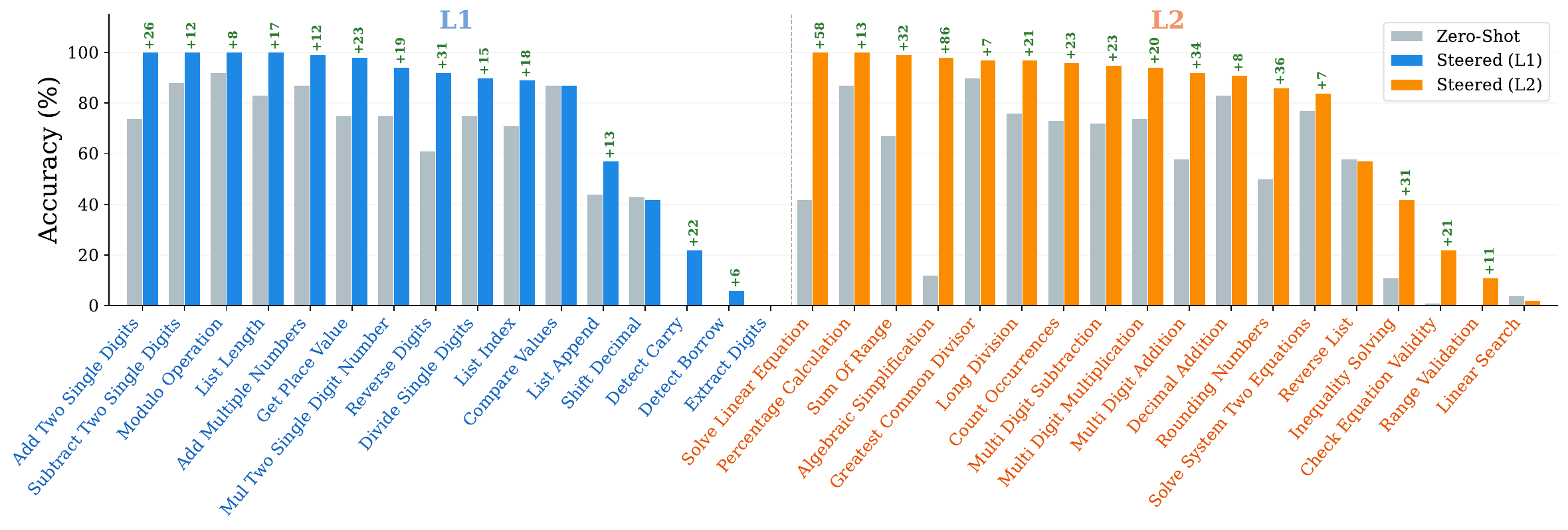}
    \caption{Per-skill accuracy on \textsc{SkillSet-Math} for Qwen3-0.6B: unsteered baseline vs.\ steered (PCA-extracted skill direction).}
    \label{fig:rq1-06b}
\end{figure*}
\Cref{fig:rq1-06b} compares per-skill accuracy on $\mathcal{D}_{\text{math}}$ between the unsteered baseline and the steered model for Qwen3-0.6B.
The PCA-extracted direction improves accuracy on 29 of 34 skills, with an average gain of 19.1 percentage points.
The largest gains appear in the composite skills, e.g., \texttt{algebraic\_simplification} ($+86$ points) and \texttt{solve\_linear\_equation} ($+58$ points), suggesting that skill-conditioned directions help activate multi-step procedures that the model fails to execute under zero-shot prompting.

\textbf{Result~2: Math-skill directions transfer to real benchmarks across model sizes.}
\Cref{tab:math_datasets_steering} reports zero-shot through 4-shot prompting and our \textbf{Steered} configuration (the math-skill direction is applied at inference time on GSM8K and MATH500) across Qwen3 models from 0.6B to 14B.
We highlight two qualitative observations.
First, few-shot prompting can be \emph{unstable} on smaller models; for example, 1-shot prompting reduces Qwen3-0.6B's GSM8K accuracy from 74.0\% to 46.7\%.
The Steered configuration does not exhibit such degradation and matches or exceeds the zero-shot baseline in 8 out of 10 (model, dataset) cells.
Second, on the larger models (4B, 8B, 14B) the Steered numbers are competitive but modest relative to the strongest prompting baseline, indicating that the absolute headroom for this kind of intervention shrinks as the underlying model becomes more capable on the benchmark.

\begin{table}[t]
\centering
\small
\setlength{\tabcolsep}{6pt}
\renewcommand{\arraystretch}{1.15}
\scalebox{0.78}{
\begin{tabular}{lcccccc}
\toprule
\textbf{Model} & \textbf{0-shot} & \textbf{1-shot} & \textbf{2-shot} & \textbf{3-shot} &  \textbf{4-shot} & \textbf{Steered} \\
\midrule
\multicolumn{7}{c}{\textbf{GSM8K}} \\
\midrule
Qwen3-0.6B & \underline{74.00} & 46.70 & 65.35 & 69.37 & 64.74 & \textbf{75.66} \\
Qwen3-1.7B & \underline{82.18} & 78.17 & 77.94 & 77.94 & 78.92 & \textbf{83.93} \\
Qwen3-4B   & \textbf{93.56} & \underline{92.42} & 90.30 & 90.37 & 91.13 & 92.34 \\
Qwen3-8B   & 95.07 & 94.77 & \textbf{95.45} & 94.84 & 95.30 & \underline{95.38} \\
Qwen3-14B  & 95.83 & 95.60 & 95.98 & \textbf{96.97} & 95.98 & \underline{96.21} \\
\midrule
\multicolumn{7}{c}{\textbf{MATH500}} \\
\midrule
Qwen3-0.6B & \textbf{67.00} & 65.20 & 63.80 & 65.40 & 51.00 & \underline{66.20}  \\
Qwen3-1.7B & \underline{72.60} & 65.80 & 67.60 & 70.40 & 68.20 & \textbf{73.80}  \\
Qwen3-4B   & \underline{89.60} & 80.80 & 84.20 & 80.60 & 81.40 & \textbf{91.00}  \\
Qwen3-8B   & 89.80 & \textbf{92.00} & 89.80 & 88.40 & 90.20 & \underline{90.60} \\
Qwen3-14B  & \underline{92.40} & 91.60 & 90.80 & 90.80 & 91.00 & \textbf{93.80} \\
\bottomrule
\end{tabular}}
\caption{Accuracy (\%) on real math benchmarks across the Qwen3 models from 0.6B to 14B. \textbf{Steered} is the model with one PCA-extracted \textsc{SkillSet-Math} skill direction added to hidden states at $\alpha = 2.0$ on the last 50\% of layers.}
\label{tab:math_datasets_steering}
\end{table}

\subsection{\textcolor{orange}{\textcircled{\scriptsize C}} Does vector-space composition outperform text-based composition?}
\label{sec:rq2}

\paragraph{Setup.}
Each composite skill is associated with a set of $k$ constituent atomic skill directions.
We approximate joint activation by composing these atomic directions with equal weights $\alpha_i = 1/k$ and re-normalizing:
$
v_{\mathrm{comp}} =
\frac{\sum_{i=1}^k \alpha_i v_i}{\left\lVert \sum_{i=1}^k \alpha_i v_i \right\rVert}.
$
We compare vector-space composition $f_\theta^{(v_{\mathrm{comp}},\alpha)}$ against:
(i) the unsteered zero-shot baseline $f_\theta^{(0,0)}$ (\textbf{Base});
(ii) \textbf{text-concise} (Txt-C), where the composite prompt is augmented with a concise text description;
(iii) \textbf{text-CoT} (Txt-CoT), where the prompt is augmented with a fuller chain-of-thought text decomposition (see~\Cref{app:composition} for prompt examples).
This comparison is the central axis of RQ2: \emph{text-based composition vs.\ vector-based composition}.
Because we do not directly estimate the interaction gain $\Gamma_{ij}$ of \Cref{sec:skill-steering}, we use an empirical proxy
$
\tilde{\Gamma} := \mathrm{Acc}(v_{\mathrm{comp}}, \alpha; \mathcal{D}) - \mathrm{Acc}_{\text{Txt-CoT}}(\mathcal{D}),
$
where $\tilde{\Gamma} > 0$ indicates that vector-space composition outperforms text-based composition.

\textbf{Result: Vector-space composition outperforms text-based composition on most composite skills.}
Vector composition achieves the highest accuracy on 13 of 18 skills (\Cref{tab:skill_composition}), improving the average from 51.9\% (Base) to 77.1\% (+25.2 points), compared to +23.8 points for Txt-C (text-concise) and +22.3 points for Txt-CoT (text-CoT).
The gains are most pronounced on procedurally complex tasks, such as \texttt{solve\_system\_two\_equations} and \texttt{inequality\_solving}.
Importantly, the composed direction $v_{\mathrm{comp}}$ is constructed only from atomic directions, with no composite skill supervision; the gain is therefore evidence that independently extracted skill directions admit vector-space recombination that text-based decomposition does not match.

\begin{table*}[t]
\centering
\small
\setlength{\tabcolsep}{4pt}
\renewcommand{\arraystretch}{1.12}
\scalebox{1.0}{
\begin{tabular}{lcccc | lcccc}
\toprule
\textbf{Skill} & \textbf{Base} & \textbf{Txt-C} & \textbf{Txt-CoT} & \textbf{Vec}
& \textbf{Skill} & \textbf{Base} & \textbf{Txt-C} & \textbf{Txt-CoT} & \textbf{Vec} \\
\midrule
\texttt{multi\_digit\_mul} & 74 & 94 & 93 & \colorbox{Mycolor-red}{96{\scriptsize$\uparrow$}}
& \texttt{gcd} & 90 & 97 & 97 & \colorbox{Mycolor-red}{99{\scriptsize$\uparrow$}} \\

\texttt{multi\_digit\_add} & 58 & 92 & 82 & \colorbox{Mycolor-red}{93{\scriptsize$\uparrow$}}
& \texttt{range\_valid} & 0 & \colorbox{Mycolor-red}{11} & \colorbox{Mycolor-red}{11} & 5{\scriptsize$\downarrow$} \\

\texttt{multi\_digit\_sub} & 72 & 95 & \colorbox{Mycolor-red}{98} & \colorbox{Mycolor-red}{98{\scriptsize=}}
& \texttt{linear\_search} & 4 & 2 & 2 & \colorbox{Mycolor-red}{7{\scriptsize$\uparrow$}} \\

\texttt{sum\_of\_range} & 67 & \colorbox{Mycolor-red}{99} & 97 & 96{\scriptsize$\downarrow$}
& \texttt{reverse\_list} & \colorbox{Mycolor-red}{58} & 57 & \colorbox{Mycolor-red}{58} & 52{\scriptsize$\downarrow$} \\

\texttt{decimal\_add} & 83 & 91 & 93 & \colorbox{Mycolor-red}{95{\scriptsize$\uparrow$}}
& \texttt{alg\_simplify} & 12 & 98 & 98 & \colorbox{Mycolor-red}{99{\scriptsize$\uparrow$}} \\

\texttt{solve\_sys\_2eq} & 77 & 84 & 86 & \colorbox{Mycolor-red}{96{\scriptsize$\uparrow$}}
& \texttt{count\_occur} & 73 & \colorbox{Mycolor-red}{96} & \colorbox{Mycolor-red}{96} & 94{\scriptsize$\downarrow$} \\

\texttt{rounding\_num} & 50 & 86 & 86 & \colorbox{Mycolor-red}{88{\scriptsize$\uparrow$}}
& \texttt{check\_eq\_valid} & 1 & 22 & 3 & \colorbox{Mycolor-red}{24{\scriptsize$\uparrow$}} \\

\texttt{long\_division} & 76 & 97 & \colorbox{Mycolor-red}{100} & \colorbox{Mycolor-red}{100{\scriptsize=}}
& \texttt{solve\_linear\_eq} & 42 & \colorbox{Mycolor-red}{100} & \colorbox{Mycolor-red}{100} & \colorbox{Mycolor-red}{100{\scriptsize=}} \\

\texttt{pct\_calc} & 87 & \colorbox{Mycolor-red}{100} & 99 & 99{\scriptsize=}
& \texttt{inequality\_solv} & 11 & 42 & 36 & \colorbox{Mycolor-red}{47{\scriptsize$\uparrow$}} \\

\midrule
\multicolumn{10}{r}{\textit{Average:} Base 51.9 \quad Txt-C 75.7 \quad Txt-CoT 74.2 \quad \textbf{Vec 77.1}} \\
\bottomrule
\end{tabular}}
\caption{
RQ2: text-based vs.\ vector-based composition on Qwen3-0.6B (accuracy \%).
\textbf{Base}: zero-shot. \textbf{Txt-C}/\textbf{Txt-CoT}: text-based composition (concise / chain-of-thought).
\textbf{Vec}: composition of atomic skill directions in vector space.
Highlighted cells indicate the best-performing method per skill.
}
\label{tab:skill_composition}
\end{table*}

\subsection{\textcolor{teal}{\textcircled{\scriptsize P}} Do skill directions enable context-conditioned personalization?}
\label{sec:rq3}

\paragraph{Setup.}
We evaluate whether preference-conditioned skill directions, estimated via contrastive hidden-state differences (\Cref{sec:method-extraction}), can induce context-dependent algorithmic behavior at inference time.
\textsc{SkillSet-Code} consists of two components:
(i) a controlled evaluation set of 36 hand-crafted tasks with paired solution strategies and situational contexts $\sigma$, yielding 72 task--situation instances; and
(ii) a contrastive training set for direction estimation, where each approach pair defines two context-conditioned distributions $\mathcal{D}_{\sigma_1}$ and $\mathcal{D}_{\sigma_2}$, each containing 200 solutions.
These distributions implicitly specify the preference-conditioned objective $\mathcal{J}_{\mathrm{opt}}(\cdot;\sigma)$ used to estimate skill directions.
We instantiate the latent skill parameterization by representing each context-conditioned skill as the expectation of hidden states,
$\Phi_\sigma := \mathbb{E}_{h \sim \mathcal{D}_\sigma}[h],$
so that the latent direction between two contexts is given by their difference,
$v = \Phi_{\sigma_1} - \Phi_{\sigma_2}.$
At inference time, the situational context $\sigma$ determines the activation direction of the skill direction, selecting either $+v$ or $-v$.
For example, in the DFS vs.\ BFS pair, a memory-constrained setting favors DFS, whereas a large-scale search setting favors BFS.
We evaluate Qwen3-0.6B and Qwen3-1.7B.

\paragraph{Evaluation.}
We measure preference alignment using GPT-4o-mini as an LLM judge (the full judge prompt is given in~\Cref{app:personalization}) that determines whether the generated code follows the prescribed algorithmic \emph{approach} (e.g., DFS vs.\ BFS), and report \emph{approach match rate}, the fraction of records where the judge confirms the prescribed approach.

\textbf{Result: Contrastive directions enable preference controllability.}
In~\Cref{tab:skillchoice24_category}, steering the LLM with the skill vector improves the match rate across categories on Qwen3-0.6B, indicating that the contrastively estimated direction $v$ generalizes to unseen task--situation instances.

\begin{table}
    \centering
    \small
    \setlength{\tabcolsep}{4pt}
    \scalebox{0.9}{
    \begin{tabular}{lrrrr}
        \toprule
        \multicolumn{1}{l}{} & \multicolumn{2}{c}{\textbf{Qwen3-0.6B}} & \multicolumn{2}{c}{\textbf{Qwen3-1.7B}} \\
        \cmidrule(lr){2-3} \cmidrule(lr){4-5}
        \textbf{Category} & \textbf{Base} & \textbf{Steered} & \textbf{Base} & \textbf{Steered} \\
        \midrule
        Graph Search & 16.7 & \textbf{33.3} & 25.0 & 16.7 \\
        DP / Recursion & 16.7 & \textbf{41.7} & 8.3 & 8.3 \\
        Data Struct.\ / String & 16.7 & \textbf{45.8} & 8.3 & \textbf{20.8} \\
        % \midrule
        % \textbf{All} & 16.7 & \textbf{41.7} & 12.5 & \textbf{16.7} \\
        \bottomrule
    \end{tabular}}
    \caption{Approach match rate (\%, controllability) by category, $\alpha=1.0$.}
    \label{tab:skillchoice24_category}
    % \vspace{-40pt}
\end{table}

\subsection{\textcolor{purple}{\textcircled{\scriptsize E}} How do skill directions evolve along optimization trajectories?}
\label{sec:rq4}

\paragraph{Setup.}
We study how a latent skill direction evolves along its optimization trajectory, where gradient-based updates provide a natural mechanism for continuously transforming the direction under a downstream objective.
Starting from a zero initialization $\Phi^{(0)}$, we optimize the skill-direction parameterization using Reference-free Preference Steering (RePS)~\citep{DBLP:WuRepSteer25Nuerips} for $T=200$ steps, yielding a trajectory $\mathcal{T}=\{\Phi^{(t)}\}_{t=0}^{T}$ and recording the corresponding direction $v_{\Phi^{(t)}}$ at each step.
At each checkpoint, we evaluate the direction using the \emph{approach match rate} (the same controllability metric defined in~\Cref{sec:rq3}), via the LLM-as-judge in~\Cref{app:evolution}.
This trajectory-based evaluation lets us characterize the evolution dynamics of skill directions: in particular, whether quality improves monotonically along optimization or peaks at intermediate states.

\textbf{Result: Optimal skill directions emerge before convergence.}
% Across all 9 approach pairs and both models, the highest in-distribution approach match rate is consistently achieved at an \emph{intermediate} checkpoint rather than at the final step~200 (\Cref{tab:evolution_summary}, \Cref{fig:evolution_trajectory}).
In 14 of the 18 (pair, model) settings, the highest in-distribution approach match rate is achieved before the final step~200 rather than at convergence.
On Qwen3-0.6B, 7 of 9 pairs peak strictly before step~200 ($^\dagger$), with early gains such as \textit{DFS vs.\ BFS} ($8\%\!\rightarrow\!24\%$ at step~20, then declining to $18\%$) and \textit{Stack vs.\ Recursion} ($4\%\!\rightarrow\!16\%$ at step~8, then falling to $10\%$).
A similar pattern holds on Qwen3-1.7B (e.g., \textit{QuadDP vs.\ Opt.} $52\%\!\rightarrow\!60\%$ at step~4, then degrading to $42\%$).
This trajectory-level non-monotonicity is consistent with the prior observations above: early updates appear to capture a broad, transferable skill direction, while continued optimization drives the direction toward task-specific features that degrade the approach match rate at convergence.
\begin{table}[t]
\centering\small\setlength{\tabcolsep}{5pt}
\scalebox{0.7}{
\begin{tabular}{l l rrr rrr}
\toprule
& & \multicolumn{3}{c}{\textbf{Qwen3-0.6B}} & \multicolumn{3}{c}{\textbf{Qwen3-1.7B}} \\
\cmidrule(lr){3-5}\cmidrule(lr){6-8}
\textbf{Cat.} & \textbf{Pair} & Base & Peak & Step & Base & Peak & Step \\
\midrule
Graph Search & DFS\,vs\,BFS & 8 & \textbf{24$^\dagger$} & 20 & 10 & 14$^\dagger$ & 5 \\
 & DFS\,vs\,UnionFind & 6 & \textbf{22} & 200 & 22 & \textbf{28$^\dagger$} & 15 \\
 & BFS\,vs\,Dijkstra & 22 & 22$^\dagger$ & 0 & 18 & 22 & 200 \\
 & BFS\,vs\,BiDirBFS & 64 & \textbf{72$^\dagger$} & 9 & 74 & \textbf{84$^\dagger$} & 30 \\
\midrule
DP / Recursion & Recursion\,vs\,DP & 0 & \textbf{24} & 200 & 2 & 6$^\dagger$ & 100 \\
 & QuadDP\,vs\,Opt. & 40 & \textbf{58$^\dagger$} & 100 & 52 & \textbf{60$^\dagger$} & 4 \\
\midrule
Data Struct. & Stack\,vs\,Recursion & 4 & \textbf{16$^\dagger$} & 8 & 42 & \textbf{54} & 200 \\
 & BruteForce\,vs\,Opt. & 10 & 12$^\dagger$ & 20 & 22 & 26$^\dagger$ & 1 \\
 & IndexScan\,vs\,TwoPtr & 10 & 14$^\dagger$ & 1 & 14 & \textbf{20$^\dagger$} & 9 \\
 \bottomrule
\end{tabular}}
\caption{%
  Approach match rate (\%, controllability) at initialization
  (\textbf{Base}, step~0) and at the best observed checkpoint
  during training (\textbf{Peak}), with the corresponding training
  step (\textbf{Step}). \textbf{Bold} = gain $\geq 5$\,pp over
  baseline. $^\dagger$ Peak occurs strictly before the final
  checkpoint (step~200), indicating that continued optimization
  \emph{degrades} performance.
}
\label{tab:evolution_summary}
\end{table}

\section{Conclusion}
In this work, we studied whether procedural skills in large language models can be \emph{represented} as directions in activation space and whether vector-space operations over these directions can express skill-level behaviors that text-based control cannot.
Through controlled experiments on \textsc{SkillSet-Math} and \textsc{SkillSet-Code}, and a real-benchmark transfer evaluation on GSM8K and MATH500 across Qwen3 0.6B--14B, we examined four observable properties of vector-space skill representations: \textbf{steerability}, \textbf{composition}, \textbf{personalization}, and \textbf{evolution}.
Empirically, we found that (i) skill-conditioned directions extracted via PCA enable direct activation of procedural behaviors on \textsc{SkillSet-Math} and transfer to real math benchmarks; (ii) independently extracted atomic directions can be composed in vector space to express composite skills, often outperforming text-based composition; (iii) contrastive directions over paired algorithmic strategies enable context-conditioned approach controllability; and (iv) directions optimized along training trajectories evolve non-monotonically, with intermediate states frequently outperforming fully converged solutions.

These findings support a representation-level view of procedural LLM skills: such skills admit a vector-space organization in the model's activations, complementing surface-level control via prompting.

\section{Limitations}
We close by noting the scope of this work. As an empirical study, our claims are bounded by the controlled \textsc{SkillSet} benchmarks and the Qwen3 models we evaluate, and several natural extensions to a fuller treatment of latent skills remain open; we highlight the most direct one here. Our composition study only considers equal-weighted, additive composition of atomic skill directions. A natural extension we leave to future work is richer arithmetic over skill directions, for example, $v(\text{multi\_digit\_add}) - v(\text{single\_digit\_add}) + v(\text{single\_digit\_mul})$, which would test whether the geometry of these directions encodes the underlying procedural relationships beyond a plain sum.

\section*{Acknowledgments}
This work is partially supported by NSF IIS-2432486.
We thank the anonymous reviewers for their constructive feedback.
AI assistants were used for writing assistance, including polishing portions of the related work, experimental discussion, and appendix text, and for generating plotting. 
All content was reviewed and verified by the authors.
The use of LLMs in dataset construction and evaluation is described in \Cref{app:skillset}.

\bibliography{main}

@inproceedings{yu2025selfimprovingagent,
  author       = {Jiongxiao Wang and
                  Qiaojing Yan and
                  Yawei Wang and
                  Yijun Tian and
                  Soumya Smruti Mishra and
                  Zhichao Xu and
                  Megha Gandhi and
                  Panpan Xu and
                  Lin Lee Cheong},
  editor       = {Maria Liakata and
                  Viviane P. Moreira and
                  Jiajun Zhang and
                  David Jurgens},
  title        = {Reinforcement Learning for Self-Improving Agent with Skill Library},
  booktitle    = {Proceedings of the 64th Annual Meeting of the Association for Computational
                  Linguistics (Volume 1: Long Papers), {ACL} 2026, San Diego, California,
                  United States, July 2-7, 2026},
  pages        = {1529--1550},
  publisher    = {Association for Computational Linguistics},
  year         = {2026},
  url          = {https://doi.org/10.18653/v1/2026.acl-long.69},
  doi          = {10.18653/V1/2026.ACL-LONG.69},
  bibsource    = {dblp computer science bibliography, https://dblp.org}
}

@article{wei2026compositionalgeneralizationllmsskill,
  author       = {Yifan Wei and
                  Li Du and
                  Xiaoyan Yu and
                  Yang Feng and
                  Angsheng Li},
  title        = {Towards Compositional Generalization of LLMs via Skill Taxonomy Guided
                  Data Synthesis},
  journal      = {CoRR},
  volume       = {abs/2601.03676},
  year         = {2026},
  url          = {https://doi.org/10.48550/arXiv.2601.03676},
  doi          = {10.48550/ARXIV.2601.03676},
  eprinttype   = {arXiv},
  eprint       = {2601.03676},
  bibsource    = {dblp computer science bibliography, https://dblp.org}
}

@misc{park2025instructskillmix,
  doi = {10.48550/ARXIV.2408.14774},
  
  url = {https://arxiv.org/abs/2408.14774},
  
  author = {Kaur, Simran and Park, Simon and Goyal, Anirudh and Arora, Sanjeev},
  
  title = {Instruct-SkillMix: A Powerful Pipeline for LLM Instruction Tuning},
  
  publisher = {arXiv},
  
  year = {2024},
  
  copyright = {arXiv.org perpetual, non-exclusive license}
}

@article{yang2025qwen3,
  author       = {Qwen Team},
  title        = {Qwen3 Technical Report},
  journal      = {CoRR},
  volume       = {abs/2505.09388},
  year         = {2025},
  url          = {https://doi.org/10.48550/arXiv.2505.09388},
  doi          = {10.48550/ARXIV.2505.09388},
  eprinttype   = {arXiv},
  eprint       = {2505.09388},
  bibsource    = {dblp computer science bibliography, https://dblp.org}
}

@article{feng2026personadynamiccompositionalinferencetime,
  author       = {Xiachong Feng and
                  Liang Zhao and
                  Weihong Zhong and
                  Yichong Huang and
                  Yuxuan Gu and
                  Lingpeng Kong and
                  Xiaocheng Feng and
                  Bing Qin},
  title        = {{PERSONA:} Dynamic and Compositional Inference-Time Personality Control
                  via Activation Vector Algebra},
  journal      = {CoRR},
  volume       = {abs/2602.15669},
  year         = {2026},
  url          = {https://doi.org/10.48550/arXiv.2602.15669},
  doi          = {10.48550/ARXIV.2602.15669},
  eprinttype   = {arXiv},
  eprint       = {2602.15669},
  bibsource    = {dblp computer science bibliography, https://dblp.org}
}

@inproceedings{pai2026billysteeringlargelanguage,
  author       = {Tsung{-}Min Pai and
                  Jui{-}I Wang and
                  Li{-}Chun Lu and
                  Shao{-}Hua Sun and
                  Hung{-}yi Lee and
                  Kai{-}Wei Chang},
  editor       = {Vera Demberg and
                  Kentaro Inui and
                  Llu{\'{\i}}s Marquez},
  title        = {{BILLY:} Steering Large Language Models via Merging Persona Vectors
                  for Creative Generation},
  booktitle    = {Proceedings of the 19th Conference of the European Chapter of the
                  Association for Computational Linguistics, {EACL} 2026 - Volume 1:
                  Long Papers, Rabat, Morocco, March 24-29, 2026},
  pages        = {7870--7915},
  publisher    = {Association for Computational Linguistics},
  year         = {2026},
  url          = {https://doi.org/10.18653/v1/2026.eacl-long.369},
  doi          = {10.18653/V1/2026.EACL-LONG.369},
  bibsource    = {dblp computer science bibliography, https://dblp.org}
}

@article{chen2025personavectorsmonitoringcontrolling,
  author       = {Runjin Chen and
                  Andy Arditi and
                  Henry Sleight and
                  Owain Evans and
                  Jack Lindsey},
  title        = {Persona Vectors: Monitoring and Controlling Character Traits in Language
                  Models},
  journal      = {CoRR},
  volume       = {abs/2507.21509},
  year         = {2025},
  url          = {https://doi.org/10.48550/arXiv.2507.21509},
  doi          = {10.48550/ARXIV.2507.21509},
  eprinttype   = {arXiv},
  eprint       = {2507.21509},
  bibsource    = {dblp computer science bibliography, https://dblp.org}
}

@article{li2026steeringvectorfieldscontextaware,
  author       = {Jiaqian Li and
                  Yanshu Li and
                  Kuan{-}Hao Huang},
  title        = {Steering Vector Fields for Context-Aware Inference-Time Control in
                  Large Language Models},
  journal      = {CoRR},
  volume       = {abs/2602.01654},
  year         = {2026},
  url          = {https://doi.org/10.48550/arXiv.2602.01654},
  doi          = {10.48550/ARXIV.2602.01654},
  eprinttype   = {arXiv},
  eprint       = {2602.01654},
  bibsource    = {dblp computer science bibliography, https://dblp.org}
}

@misc{yang2025automatedskilldiscoverylanguage,
      title={Automated Skill Discovery for Language Agents through Exploration and Iterative Feedback}, 
      author={Yongjin Yang and Sinjae Kang and Juyong Lee and Dongjun Lee and Se-Young Yun and Kimin Lee},
      year={2025},
      eprint={2506.04287},
      archivePrefix={arXiv},
      primaryClass={cs.AI},
      url={https://arxiv.org/abs/2506.04287}, 
}

@article{wang2025inducingprogrammaticskills,
  author       = {Zora Zhiruo Wang and
                  Apurva Gandhi and
                  Graham Neubig and
                  Daniel Fried},
  title        = {Inducing Programmatic Skills for Agentic Tasks},
  journal      = {CoRR},
  volume       = {abs/2504.06821},
  year         = {2025},
  url          = {https://doi.org/10.48550/arXiv.2504.06821},
  doi          = {10.48550/ARXIV.2504.06821},
  eprinttype   = {arXiv},
  eprint       = {2504.06821},
  bibsource    = {dblp computer science bibliography, https://dblp.org}
}

@article{xia2026skillrl,
  author       = {Peng Xia and
                  Jianwen Chen and
                  Hanyang Wang and
                  Jiaqi Liu and
                  Kaide Zeng and
                  Yu Wang and
                  Siwei Han and
                  Yiyang Zhou and
                  Xujiang Zhao and
                  Haifeng Chen and
                  Zeyu Zheng and
                  Cihang Xie and
                  Huaxiu Yao},
  title        = {SkillRL: Evolving Agents via Recursive Skill-Augmented Reinforcement
                  Learning},
  journal      = {CoRR},
  volume       = {abs/2602.08234},
  year         = {2026},
  url          = {https://doi.org/10.48550/arXiv.2602.08234},
  doi          = {10.48550/ARXIV.2602.08234},
  eprinttype   = {arXiv},
  eprint       = {2602.08234},
  bibsource    = {dblp computer science bibliography, https://dblp.org}
}

@article{representation_eng,
  author       = {Andy Zou and
                  Long Phan and
                  Sarah Li Chen and
                  James Campbell and
                  Phillip Guo and
                  Richard Ren and
                  Alexander Pan and
                  Xuwang Yin and
                  Mantas Mazeika and
                  Ann{-}Kathrin Dombrowski and
                  Shashwat Goel and
                  Nathaniel Li and
                  Michael J. Byun and
                  Zifan Wang and
                  Alex Mallen and
                  Steven Basart and
                  Sanmi Koyejo and
                  Dawn Song and
                  Matt Fredrikson and
                  J. Zico Kolter and
                  Dan Hendrycks},
  title        = {Representation Engineering: {A} Top-Down Approach to {AI} Transparency},
  journal      = {CoRR},
  volume       = {abs/2310.01405},
  year         = {2023},
  url          = {https://doi.org/10.48550/arXiv.2310.01405},
  doi          = {10.48550/ARXIV.2310.01405},
  eprinttype   = {arXiv},
  eprint       = {2310.01405},
  bibsource    = {dblp computer science bibliography, https://dblp.org}
}

@article{zheng2025skillweaver,
  author       = {Boyuan Zheng and
                  Michael Y. Fatemi and
                  Xiaolong Jin and
                  Zora Zhiruo Wang and
                  Apurva Gandhi and
                  Yueqi Song and
                  Yu Gu and
                  Jayanth Srinivasa and
                  Gaowen Liu and
                  Graham Neubig and
                  Yu Su},
  title        = {SkillWeaver: Web Agents can Self-Improve by Discovering and Honing
                  Skills},
  journal      = {CoRR},
  volume       = {abs/2504.07079},
  year         = {2025},
  url          = {https://doi.org/10.48550/arXiv.2504.07079},
  doi          = {10.48550/ARXIV.2504.07079},
  eprinttype   = {arXiv},
  eprint       = {2504.07079},
  bibsource    = {dblp computer science bibliography, https://dblp.org}
}

@article{zhang2026memskill,
  title={MemSkill: Learning and Evolving Memory Skills for Self-Evolving Agents},
  author={Zhang, Haozhen and Long, Quanyu and Bao, Jianzhu and Feng, Tao and Zhang, Weizhi and Yue, Haodong and Wang, Wenya},
  journal={arXiv preprint arXiv:2602.02474},
  year={2026}
}

@inproceedings{DBLP:conf/nips/CaoZC00MC24,
  author       = {Yuanpu Cao and
                  Tianrong Zhang and
                  Bochuan Cao and
                  Ziyi Yin and
                  Lu Lin and
                  Fenglong Ma and
                  Jinghui Chen},
  editor       = {Amir Globersons and
                  Lester Mackey and
                  Danielle Belgrave and
                  Angela Fan and
                  Ulrich Paquet and
                  Jakub M. Tomczak and
                  Cheng Zhang},
  title        = {Personalized Steering of Large Language Models: Versatile Steering
                  Vectors Through Bi-directional Preference Optimization},
  booktitle    = {Advances in Neural Information Processing Systems 38: Annual Conference
                  on Neural Information Processing Systems 2024, NeurIPS 2024, Vancouver,
                  BC, Canada, December 10 - 15, 2024},
  year         = {2024},
  url          = {http://papers.nips.cc/paper\_files/paper/2024/hash/58cbe393b4254da8966780a40d023c0b-Abstract-Conference.html},
  bibsource    = {dblp computer science bibliography, https://dblp.org}
}

@inproceedings{DBLP:conf/nips/ArditiOSPPGN24,
  author       = {Andy Arditi and
                  Oscar Obeso and
                  Aaquib Syed and
                  Daniel Paleka and
                  Nina Panickssery and
                  Wes Gurnee and
                  Neel Nanda},
  editor       = {Amir Globersons and
                  Lester Mackey and
                  Danielle Belgrave and
                  Angela Fan and
                  Ulrich Paquet and
                  Jakub M. Tomczak and
                  Cheng Zhang},
  title        = {Refusal in Language Models Is Mediated by a Single Direction},
  booktitle    = {Advances in Neural Information Processing Systems 38: Annual Conference
                  on Neural Information Processing Systems 2024, NeurIPS 2024, Vancouver,
                  BC, Canada, December 10 - 15, 2024},
  year         = {2024},
  url          = {http://papers.nips.cc/paper\_files/paper/2024/hash/f545448535dfde4f9786555403ab7c49-Abstract-Conference.html},
  bibsource    = {dblp computer science bibliography, https://dblp.org}
}

@article{DBLP:Tonebank,
  title={Beyond linear steering: Unified multi-attribute control for language models},
  author={Oozeer, Narmeen and Marks, Luke and Barez, Fazl and Abdullah, Amirali},
  journal={Findings of the Association for Computational Linguistics: EMNLP},
  pages={23513--23557},
  year={2025}
}

@article{DBLP:WuRepSteer25Nuerips,
  author       = {Zhengxuan Wu and
                  Qinan Yu and
                  Aryaman Arora and
                  Christopher D. Manning and
                  Christopher Potts},
  title        = {Improved Representation Steering for Language Models},
  journal      = {CoRR},
  volume       = {abs/2505.20809},
  year         = {2025},
  url          = {https://doi.org/10.48550/arXiv.2505.20809},
  doi          = {10.48550/ARXIV.2505.20809},
  eprinttype    = {arXiv},
  eprint       = {2505.20809},
  bibsource    = {dblp computer science bibliography, https://dblp.org}
}

@inproceedings{DBLP:conf/nips/WuAWGJMP24,
  author       = {Zhengxuan Wu and
                  Aryaman Arora and
                  Zheng Wang and
                  Atticus Geiger and
                  Dan Jurafsky and
                  Christopher D. Manning and
                  Christopher Potts},
  editor       = {Amir Globersons and
                  Lester Mackey and
                  Danielle Belgrave and
                  Angela Fan and
                  Ulrich Paquet and
                  Jakub M. Tomczak and
                  Cheng Zhang},
  title        = {ReFT: Representation Finetuning for Language Models},
  booktitle    = {Advances in Neural Information Processing Systems 38: Annual Conference
                  on Neural Information Processing Systems 2024, NeurIPS 2024, Vancouver,
                  BC, Canada, December 10 - 15, 2024},
  year         = {2024},
  url          = {http://papers.nips.cc/paper\_files/paper/2024/hash/75008a0fba53bf13b0bb3b7bff986e0e-Abstract-Conference.html},
  bibsource    = {dblp computer science bibliography, https://dblp.org}
}

@inproceedings{DBLP:conf/iclr/LeePRMDND25,
  author       = {Bruce W. Lee and
                  Inkit Padhi and
                  Karthikeyan Natesan Ramamurthy and
                  Erik Miehling and
                  Pierre L. Dognin and
                  Manish Nagireddy and
                  Amit Dhurandhar},
  title        = {Programming Refusal with Conditional Activation Steering},
  booktitle    = {The Thirteenth International Conference on Learning Representations,
                  {ICLR} 2025, Singapore, April 24-28, 2025},
  publisher    = {OpenReview.net},
  year         = {2025},
  url          = {https://openreview.net/forum?id=Oi47wc10sm},
  bibsource    = {dblp computer science bibliography, https://dblp.org}
}

@inproceedings{DBLP:SAKE,
  author       = {Marco Scialanga and
                  Thibault Laugel and
                  Vincent Grari and
                  Marcin Detyniecki},
  editor       = {Wanxiang Che and
                  Joyce Nabende and
                  Ekaterina Shutova and
                  Mohammad Taher Pilehvar},
  title        = {{SAKE:} Steering Activations for Knowledge Editing},
  booktitle    = {Proceedings of the 63rd Annual Meeting of the Association for Computational
                  Linguistics (Volume 1: Long Papers), {ACL} 2025, Vienna, Austria,
                  July 27 - August 1, 2025},
  pages        = {15966--15978},
  publisher    = {Association for Computational Linguistics},
  year         = {2025},
  url          = {https://aclanthology.org/2025.acl-long.777/},
  bibsource    = {dblp computer science bibliography, https://dblp.org}
}

@article{easysteer,
  author       = {Haolei Xu and
                  Xinyu Mei and
                  Yuchen Yan and
                  Rui Zhou and
                  Wenqi Zhang and
                  Weiming Lu and
                  Yueting Zhuang and
                  Yongliang Shen},
  title        = {EasySteer: {A} Unified Framework for High-Performance and Extensible
                  {LLM} Steering},
  journal      = {CoRR},
  volume       = {abs/2509.25175},
  year         = {2025},
  url          = {https://doi.org/10.48550/arXiv.2509.25175},
  doi          = {10.48550/ARXIV.2509.25175},
  eprinttype    = {arXiv},
  eprint       = {2509.25175},
  bibsource    = {dblp computer science bibliography, https://dblp.org}
}

@article{SKiC,
  author       = {Jiaao Chen and
                  Xiaoman Pan and
                  Dian Yu and
                  Kaiqiang Song and
                  Xiaoyang Wang and
                  Dong Yu and
                  Jianshu Chen},
  title        = {Skills-in-Context Prompting: Unlocking Compositionality in Large Language
                  Models},
  journal      = {CoRR},
  volume       = {abs/2308.00304},
  year         = {2023},
  url          = {https://doi.org/10.48550/arXiv.2308.00304},
  doi          = {10.48550/ARXIV.2308.00304},
  eprinttype    = {arXiv},
  eprint       = {2308.00304},
  bibsource    = {dblp computer science bibliography, https://dblp.org}
}

@misc{cho2025corrsteergenerationtimellmsteering,
      title={CorrSteer: Generation-Time LLM Steering via Correlated Sparse Autoencoder Features}, 
      author={Seonglae Cho and Zekun Wu and Adriano Koshiyama},
      year={2025},
      eprint={2508.12535},
      archivePrefix={arXiv},
      primaryClass={cs.CL},
      url={https://arxiv.org/abs/2508.12535}, 
}

@article{ComposableChainsofThought,
  author       = {Fangcong Yin and
                  Zeyu Leo Liu and
                  Liu Leqi and
                  Xi Ye and
                  Greg Durrett},
  title        = {Learning Composable Chains-of-Thought},
  journal      = {CoRR},
  volume       = {abs/2505.22635},
  year         = {2025},
  url          = {https://doi.org/10.48550/arXiv.2505.22635},
  doi          = {10.48550/ARXIV.2505.22635},
  eprinttype    = {arXiv},
  eprint       = {2505.22635},
  bibsource    = {dblp computer science bibliography, https://dblp.org}
}

@misc{park2025doesrlposttraininginduce,
      title={How Does RL Post-training Induce Skill Composition? A Case Study on Countdown}, 
      author={Simon Park and Simran Kaur and Sanjeev Arora},
      year={2025},
      eprint={2512.01775},
      archivePrefix={arXiv},
      primaryClass={cs.LG},
      url={https://arxiv.org/abs/2512.01775}, 
}

@article{alzubi2026evoskill,
  title={EvoSkill: Automated Skill Discovery for Multi-Agent Systems},
  author={Alzubi, Salaheddin and Provenzano, Noah and Bingham, Jaydon and Chen, Weiyuan and Vu, Tu},
  journal={arXiv preprint arXiv:2603.02766},
  year={2026}
}

@inproceedings{subramani2022extracting,
  title={Extracting latent steering vectors from pretrained language models},
  author={Subramani, Nishant and Suresh, Nivedita and Peters, Matthew E},
  booktitle={Findings of the Association for Computational Linguistics: ACL 2022},
  pages={566--581},
  year={2022}
}

@article{turner2023steering,
  title={Steering language models with activation engineering},
  author={Turner, Alexander Matt and Thiergart, Lisa and Leech, Gavin and Udell, David and Vazquez, Juan J and Mini, Ulisse and MacDiarmid, Monte},
  journal={arXiv preprint arXiv:2308.10248},
  year={2023}
}

@article{ali2025scaling,
  author       = {Sheikh Abdur Raheem Ali and
                  Justin Xu and
                  Ivory Yang and
                  Jasmine Xinze Li and
                  Ayse Arslan and
                  Clark Benham},
  title        = {Scaling laws for activation steering with Llama 2 models and refusal
                  mechanisms},
  journal      = {CoRR},
  volume       = {abs/2507.11771},
  year         = {2025},
  url          = {https://doi.org/10.48550/arXiv.2507.11771},
  doi          = {10.48550/ARXIV.2507.11771},
  eprinttype   = {arXiv},
  eprint       = {2507.11771},
  bibsource    = {dblp computer science bibliography, https://dblp.org}
}

@inproceedings{panickssery2023steering,
    title = "Steering Llama 2 via Contrastive Activation Addition",
    author = "Rimsky, Nina  and
      Gabrieli, Nick  and
      Schulz, Julian  and
      Tong, Meg  and
      Hubinger, Evan  and
      Turner, Alexander",
    editor = "Ku, Lun-Wei  and
      Martins, Andre  and
      Srikumar, Vivek",
    booktitle = "Proceedings of the 62nd Annual Meeting of the Association for Computational Linguistics (Volume 1: Long Papers)",
    month = aug,
    year = "2024",
    address = "Bangkok, Thailand",
    publisher = "Association for Computational Linguistics",
    url = "https://aclanthology.org/2024.acl-long.828/",
    doi = "10.18653/v1/2024.acl-long.828",
    pages = "15504--15522"
}

@article{atzmon2020causal,
  title={A causal view of compositional zero-shot recognition},
  author={Atzmon, Yuval and Kreuk, Felix and Shalit, Uri and Chechik, Gal},
  journal={Advances in Neural Information Processing Systems},
  volume={33},
  pages={1462--1473},
  year={2020}
}

@inproceedings{naeem2021learning,
  title={Learning graph embeddings for compositional zero-shot learning},
  author={Naeem, Muhammad Ferjad and Xian, Yongqin and Tombari, Federico and Akata, Zeynep},
  booktitle={Proceedings of the IEEE/CVF conference on computer vision and pattern recognition},
  pages={953--962},
  year={2021}
}

@inproceedings{zheng2021deep,
  title={Deep compositional metric learning},
  author={Zheng, Wenzhao and Wang, Chengkun and Lu, Jiwen and Zhou, Jie},
  booktitle={Proceedings of the IEEE/CVF conference on computer vision and pattern recognition},
  pages={9320--9329},
  year={2021}
}

@article{kaya2019deep,
  title={Deep metric learning: A survey},
  author={Kaya, Mahmut and Bilge, Hasan {\c{S}}akir},
  journal={Symmetry},
  volume={11},
  number={9},
  pages={1066},
  year={2019},
  publisher={MDPI}
}
\appendix

\section{SkillSet Construction}
\label{app:skillset}

\textsc{SkillSet} consists of two domains: math and code, designed for controlled evaluation of latent skill properties.

\subsection{Programmatic Synthesis}

We construct \textsc{SkillSet} as a controlled benchmark for skill-level evaluation.
In the current setup, \textsc{SkillSet-Math} contains 16 atomic skills and 18 composite skills. Composite skills are compositions of atomic skills as shown in Table~\ref{tab:skill-inventory}.
The construction of SkillSet can be divided into three parts: the instruction for each skill, the question, and the answer.
For composite skills, we use deterministic code to generate the answer.
Each skill contains 1000 questions, split into training, validation, and testing sets with an 8:1:1 ratio.
In later sections, we use the training data to construct steering vectors and the testing set to evaluate the effectiveness of steering.

\paragraph{SkillSet-Math}
\label{app:skillset-math}
We construct \textbf{SkillSet-Math} as a hierarchical math-reasoning benchmark with two levels.

\textbf{Atomic} contains 16 atomic skills, including single-digit arithmetic (addition, subtraction, multiplication, division), digit manipulation (extraction, reversal), carry/borrow detection, list primitives (indexing, length, append), and value comparison.  
\textbf{Composite} contains 18 composite skills, covering multi-digit arithmetic (addition, subtraction, multiplication, long division), decimal and percentage computation, algebraic tasks (linear equations, systems of two equations, inequality solving, simplification), number-theoretic routines (GCD, range validation, rounding), and list algorithms (linear search, reverse, count occurrences).  
Each composite skill is explicitly defined as a composition of 2--4 atomic sub-skills (Table~\ref{tab:l2-deps}).

The dependency structure is explicit. For example,
\texttt{multi\_digit\_addition} decomposes into \texttt{extract\_digits}
$\to$ \texttt{add\_two\_single\_digits} $\to$ \texttt{detect\_carry}
$\to$ \texttt{reverse\_digits}, and \texttt{solve\_system\_two\_equations}
decomposes into \texttt{mul\_two\_single\_digit\_number} $\to$
\texttt{subtract\_two\_single\_digits} $\to$
\texttt{divide\_single\_digits}.

This compositional hierarchy enables our core experiment: we extract one steering vector per atomic skill from 200 training examples, and compose multiple atomic vectors at inference time to steer composite task performance without any composite skill training. 
This setup tests whether independently learned skill vectors combine to produce emergent compositional capabilities.
Table~\ref{tab:skill-inventory} lists the full inventory, and Table~\ref{tab:l2-deps} provides the dependency mapping.
Examples of atomic and composite skills are shown in Figure~\ref{skillcard:l1}

\begin{table}[htbp]
\centering
\small
\setlength{\tabcolsep}{6pt}
\renewcommand{\arraystretch}{1.15}

\begin{tabularx}{\linewidth}{@{}p{0.12\linewidth} c X@{}}
\toprule
\textbf{Level} & \textbf{\#} & \textbf{Skills} \\
\midrule
atomic & 16 &
\texttt{extract\_digits},
\texttt{reverse\_digits},
\texttt{add\_two\_single\_digits},
\texttt{subtract\_two\_single\_digits},
\texttt{mul\_two\_single\_digit\_number},
\texttt{divide\_single\_digits},
\texttt{add\_multiple\_numbers},
\texttt{compare\_values},
\texttt{detect\_carry},
\texttt{detect\_borrow},
\texttt{modulo\_operation},
\texttt{get\_place\_value},
\texttt{shift\_decimal},
\texttt{list\_index},
\texttt{list\_length},
\texttt{list\_append}
\\ \midrule
composite & 18 &
\texttt{multi\_digit\_addition},
\texttt{multi\_digit\_subtraction},
\texttt{multi\_digit\_multiplication},
\texttt{long\_division},
\texttt{decimal\_addition},
\texttt{sum\_of\_range},
\texttt{rounding\_numbers},
\texttt{percentage\_calculation},
\texttt{greatest\_common\_divisor},
\texttt{solve\_linear\_equation},
\texttt{solve\_system\_two\_equations},
\texttt{check\_equation\_validity},
\texttt{inequality\_solving},
\texttt{algebraic\_simplification},
\texttt{range\_validation},
\texttt{linear\_search},
\texttt{reverse\_list},
\texttt{count\_occurrences}
\\
\bottomrule
\end{tabularx}

\caption{Skill inventory used in our experiments.
  atomic skills are atomic operations (single-digit arithmetic, list
  manipulation, comparisons); composite skills are composite procedures that
  each depend on 2--4 atomic sub-skills.}
\label{tab:skill-inventory}
\end{table}

\begin{skillcard}{Atomic Skill Example: add\_multiple\_numbers}
\small
\textbf{Instruction:} Add all numbers in a list together sequentially. Output type: int. Example: Add all numbers in this list: [1, 2, 3, 4]. Answer: 10

\renewcommand{\arraystretch}{1.25}
\begin{tabularx}{\linewidth}{>{\raggedright\arraybackslash}X
                            >{\raggedright\arraybackslash}p{0.18\linewidth}}
\textbf{Question} & \textbf{Answer} \\
\toprule
Add all numbers in this list: [1, 2, 3, 4]. & 10 \\
Add all numbers in this list: [5, 1]. & 6 \\
Add all numbers in this list: [0, 9, 9]. & 18 \\
\end{tabularx}
\label{skillcard:l1}
\end{skillcard}

\begin{skillcard}{Composite Skill Example: algebraic\_simplification}
\small
\textbf{Instruction:} Simplify an algebraic expression by combining like terms (format: coefficient*x + constant). Output type: string. Steps: 1) Identify like terms (same variable and power), 2) Add coefficients of like terms, 3) Combine constants. Your final answer must be a simplified expression (e.g., ``21x+13'' or ``5x-3'').
\renewcommand{\arraystretch}{1.25}
\begin{tabularx}{\linewidth}{>{\raggedright\arraybackslash}X
                            >{\raggedright\arraybackslash}p{0.6\linewidth}}
\textbf{Question} & \textbf{Answer} \\
\toprule
Simplify 1x+4x & 1. Using Skill \texttt{<compare\_values>}, identify like terms: x-terms=[`1x', `4x'], constants=[]. 2. Using Skill \texttt{<add\_multiple\_numbers>}, combine x-terms: 1+4=5. 3. So the answer is 5x. \\
\end{tabularx}
\label{skillcard:l2}
\end{skillcard}

\begin{table*}[t]
  \centering
  \small
  \caption{Composite skill dependencies. Each composite skill is composed of 2--4
  atomic sub-skills applied in sequence.}
  \label{tab:l2-deps}
  \setlength{\tabcolsep}{4pt}
  \renewcommand{\arraystretch}{1.1}
  \begin{tabularx}{\linewidth}{@{}p{0.32\linewidth} X@{}}
  \toprule
  \textbf{Composite Skill} & \textbf{Atomic skill Dependencies} \\
  \midrule
  \texttt{multi\_digit\_addition}
    & \texttt{extract\_digits}, \texttt{add\_two\_single\_digits},
      \texttt{detect\_carry}, \texttt{reverse\_digits} \\
  \texttt{multi\_digit\_subtraction}
    & \texttt{extract\_digits}, \texttt{subtract\_two\_single\_digits},
      \texttt{detect\_borrow}, \texttt{reverse\_digits} \\
  \texttt{multi\_digit\_multiplication}
    & \texttt{extract\_digits}, \texttt{mul\_two\_single\_digit\_number},
      \texttt{add\_multiple\_numbers}, \texttt{shift\_decimal} \\
  \texttt{long\_division}
    & \texttt{extract\_digits}, \texttt{mul\_two\_single\_digit\_number},
      \texttt{subtract\_two\_single\_digits}, \texttt{compare\_values} \\
  \texttt{decimal\_addition}
    & \texttt{extract\_digits}, \texttt{add\_two\_single\_digits},
      \texttt{detect\_carry}, \texttt{get\_place\_value} \\
  \texttt{sum\_of\_range}
    & \texttt{add\_multiple\_numbers}, \texttt{compare\_values},
      \texttt{subtract\_two\_single\_digits} \\
  \texttt{rounding\_numbers}
    & \texttt{extract\_digits}, \texttt{get\_place\_value},
      \texttt{compare\_values} \\
  \texttt{percentage\_calculation}
    & \texttt{mul\_two\_single\_digit\_number},
      \texttt{divide\_single\_digits}, \texttt{shift\_decimal} \\
  \texttt{greatest\_common\_divisor}
    & \texttt{modulo\_operation}, \texttt{compare\_values} \\
  \texttt{solve\_linear\_equation}
    & \texttt{subtract\_two\_single\_digits},
      \texttt{divide\_single\_digits}, \texttt{compare\_values} \\
  \texttt{solve\_system\_two\_equations}
    & \texttt{mul\_two\_single\_digit\_number},
      \texttt{subtract\_two\_single\_digits},
      \texttt{divide\_single\_digits} \\
  \texttt{check\_equation\_validity}
    & \texttt{mul\_two\_single\_digit\_number},
      \texttt{add\_multiple\_numbers}, \texttt{compare\_values} \\
  \texttt{inequality\_solving}
    & \texttt{subtract\_two\_single\_digits},
      \texttt{divide\_single\_digits}, \texttt{compare\_values} \\
  \texttt{algebraic\_simplification}
    & \texttt{add\_multiple\_numbers},
      \texttt{mul\_two\_single\_digit\_number},
      \texttt{compare\_values} \\
  \texttt{range\_validation}
    & \texttt{compare\_values} \\
  \texttt{linear\_search}
    & \texttt{list\_index}, \texttt{compare\_values},
      \texttt{list\_length} \\
  \texttt{reverse\_list}
    & \texttt{list\_index}, \texttt{list\_length},
      \texttt{list\_append} \\
  \texttt{count\_occurrences}
    & \texttt{list\_index}, \texttt{compare\_values},
      \texttt{list\_length} \\
  \bottomrule
  \end{tabularx}
\end{table*}

\subsection{LLM-Based Data Generation.}
\begin{figure*}
    \centering
    \includegraphics[width=\linewidth]{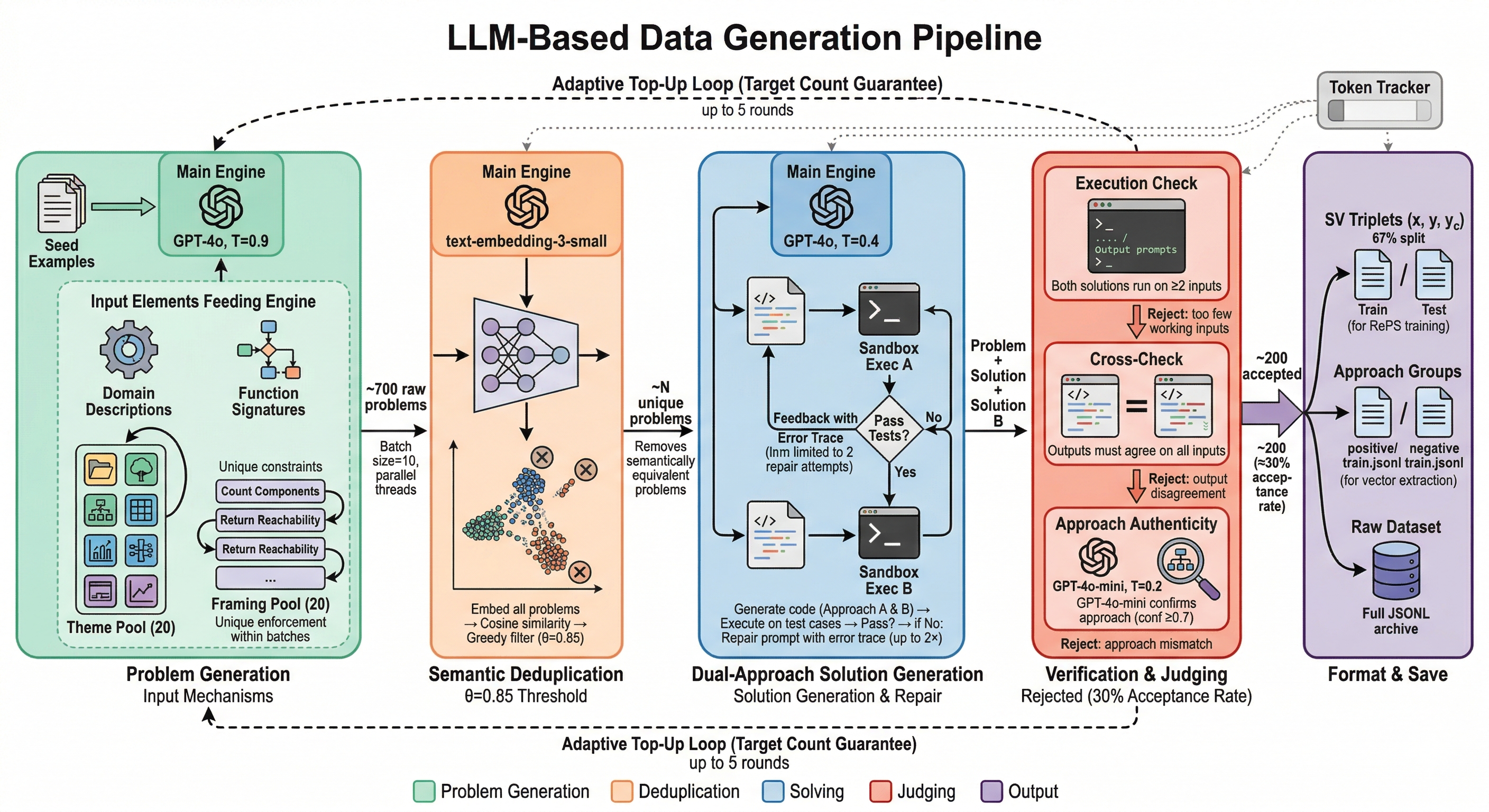}
    \caption{LLM-Based Data Generation}
    \label{fig:data_gen_pipeline}
\end{figure*}
To construct diverse, verified problem-solution datasets for each approach pair
(e.g., DFS vs.\ BFS), we design a multi-stage generation pipeline driven by
large language models, as shown in Figure~\ref{fig:data_gen_pipeline}. 
First, \textbf{GPT-4o} (temperature $0.9$) generates
novel coding problems in batches, guided by pair-specific domain descriptions,
function signature templates, and two diversity mechanisms: a pool of 20
\emph{thematic contexts} (e.g., grid traversal, tree search) assigned
round-robin across batches, and a pool of 20 \emph{output framings} (e.g.,
count components, return reachability) enforced to be unique within each batch.
We generate $3.5\times$ the target count to absorb downstream attrition. The
raw problems are then semantically deduplicated using cosine
similarity over \texttt{text-embedding-3-small} embeddings with a threshold of
$0.85$. Next, a \textbf{solution generator} (GPT-4o, temperature $0.4$)
produces two solutions per problem---one for each approach---using a self-test
and repair loop: each candidate solution is executed against the problem's test
cases in a sandboxed subprocess, and if any test fails, the error trace is fed
back to the LLM for up to two repair attempts. The resulting solution pairs
undergo a \textbf{three-stage verification}: (i)~\emph{execution
verification}, requiring both solutions to run successfully on at least two
test inputs; (ii)~\emph{cross-check agreement}, requiring identical outputs on
all mutually successful inputs; and (iii)~\emph{approach authenticity
verification}, where \textbf{GPT-4o-mini} (temperature $0.2$) inspects each
solution's code structure and confirms the claimed algorithmic approach with
confidence $\geq 0.7$. An adaptive top-up loop re-generates problems for up to
five rounds to guarantee the target count (default 200) is met. The accepted
items are formatted as contrastive triplets $(x, y, y_c)$---where $x$ is the
problem, $y$ is the approach to steer away from, and $y_c$ is the approach to
steer toward---and split into train/test sets (67\%/33\%) for downstream
steering vector extraction and evaluation.

\paragraph{\textsc{SkillSet-Code} Dataset}
\label{app:skillset-code}

The \textsc{SkillSet-Code} dataset is designed to evaluate two capabilities:
\textbf{(RQ3)\textcolor{teal}{\textcircled{\scriptsize P}}}~situational personalization, where the model selects among
algorithmic approaches based on context (e.g., memory constraints, input scale,
readability preference), and \textbf{(RQ4) \textcolor{purple}{\textcircled{\scriptsize E}}}~skill evolution, where a model's
default approach preference is shifted along a contrastive trajectory via
Reference-free Preference Steering (RePS). Both research questions share the
same underlying data but consume it in different formats.

\paragraph{Approach Pairs and Steering Vectors.}
We define 9 contrastive approach pairs (Table~\ref{tab:pairs-short}), each
representing two algorithmically valid strategies for a common problem domain
(e.g., DFS vs.\ BFS for graph traversal, recursion with memoization vs.\
bottom-up DP for optimization). For each pair, a steering vector is computed as
$v = \mathrm{mean}(h^{+}) - \mathrm{mean}(h^{-})$
over hidden-state activations from 200 positive-approach and 200
negative-approach solutions, enabling fine-grained control over which strategy
the model favors.

\noindent
The final acceptance rate is approximately 30\%, yielding 200 verified pairs per
approach pair at a total API cost of ${\sim}\$63$.

\paragraph{Data Formats.}
The accepted items are exported in two formats corresponding to the two
research questions:

\begin{itemize}
  \item \textbf{Approach groups} (for RQ3 \textcolor{teal}{\textcircled{\scriptsize P}}, personalization).
    Each pair produces \texttt{positive/train.jsonl} and
    \texttt{negative/train.jsonl}, where every entry is formatted as
    \texttt{"Problem: \ldots\textbackslash n\textbackslash nSolution
    (Approach):\textbackslash n<code>"}. Steering vectors are extracted from
    the hidden-state contrast between positive and negative groups.
    At inference time, a \emph{situational prompt} determines the steering
    direction: e.g., a memory-constrained environment steers toward DFS,
    while a large-grid setting steers toward BFS
    (Table~\ref{tab:situations}).

  \item \textbf{Contrastive triplets} (for RQ4  \textcolor{purple}{\textcircled{\scriptsize E}}, skill evolution).
    Each problem yields a triplet $(x, y, y_c)$ where $x$ is the problem
    statement, $y$ is the solution using the approach to steer \emph{away}
    from, and $y_c$ is the solution using the approach to steer \emph{toward}.
    Triplets are split 67\%/33\% into train/test sets with no overlapping
    problems. These triplets define the contrastive trajectory for RePS
    training, enabling the model's default approach preference to be
    progressively shifted.
\end{itemize}

\begin{table*}[t]
  \centering
  \caption{The 9 contrastive approach pairs in \textsc{SkillSet-Code}.}
  \label{tab:pairs-short}
  \small
  \begin{tabular}{clll}
  \toprule
  \# & \textbf{Pair} & \textbf{Approach A} & \textbf{Approach B} \\
  \midrule
  1 & \texttt{dfs\_vs\_bfs}               & DFS                  & BFS \\
  2 & \texttt{recursion\_vs\_dp}           & Recursion + Memo     & Bottom-Up DP \\
  3 & \texttt{brute\_force\_vs\_optimized} & Brute Force $O(n^2)$ & Optimized $O(n)$ \\
  4 & \texttt{stack\_vs\_recursion}        & Explicit Stack       & Recursive Descent \\
  5 & \texttt{bfs\_vs\_dijkstra}           & BFS (unweighted)     & Dijkstra (weighted) \\
  6 & \texttt{dfs\_vs\_union\_find}        & DFS traversal        & Union-Find \\
  7 & \texttt{bfs\_vs\_bidirectional\_bfs} & Standard BFS         & Bidirectional BFS \\
  8 & \texttt{quadratic\_dp\_vs\_optimized}& $O(n^2)$ DP          & $O(n\log n)$ Binary Search \\
  9 & \texttt{index\_scan\_vs\_two\_pointer}& Index Scan           & Two-Pointer \\
  \bottomrule
  \end{tabular}
\end{table*}

\begin{table*}[t]
\centering
\caption{Overview of the \textsc{SkillSet-Code} evaluation dataset. Each task presents two alternative solution methods and two situational contexts that each favor one method.}
\label{tab:situations}
\small
\begin{tabular}{@{}llll@{}}
\toprule
\textbf{Category} & \textbf{Task} & \textbf{Method A} & \textbf{Method B} \\
\midrule
\multirow{6}{*}{\shortstack[l]{Graph Search\\(6 tasks)}}
  & Number of Islands       & DFS              & BFS \\
  & Connected Components    & DFS              & Union-Find \\
  & Shortest Path (Grid)    & BFS              & Dijkstra \\
  & Word Ladder             & BFS              & Bidirectional BFS \\
  & Course Schedule         & DFS              & BFS (Kahn's) \\
  & Bipartite Check         & DFS              & BFS \\
\midrule
\multirow{6}{*}{\shortstack[l]{DP \& Recursion\\(6 tasks)}}
  & Climbing Stairs         & Memoization      & Iterative DP \\
  & Coin Change             & Memoization      & Bottom-Up DP \\
  & Decode Ways             & Memoization      & Bottom-Up DP \\
  & Longest Increasing Subseq. & $O(n^2)$ DP  & $O(n\log n)$ Binary Search \\
  & Edit Distance           & Memoization      & Bottom-Up DP \\
  & Partition Equal Subset   & Memoization      & 0/1 Knapsack DP \\
\midrule
\multirow{6}{*}{\shortstack[l]{DS \& String\\(6 tasks)}}
  & Valid Parentheses       & Stack            & Counter \\
  & Decode String           & Stack            & Recursion \\
  & Evaluate Expression     & Stack Parsing    & Recursive Descent \\
  & Run-Length Encoding     & Index Scan       & Two-Pointer \\
  & Basic Calculator        & Stack            & Recursion \\
  & Next Greater Element    & Brute Force      & Monotonic Stack \\
\bottomrule
\end{tabular}
\vspace{0.5em}
\end{table*}

\section{Experiments Details}

\subsection{Steerability}

\paragraph{Prompt} As for prompt during the inference, we require LLM to write the answer in \textbackslash boxed\{\} to help proper evaluation. In MATH500, we follow the prompt in EasySteer \citep{easysteer}.
\begin{promptboxmath}{MATH500 inference prompt}
Please reason step by step, and put your final answer within \textbackslash boxed\{\}.

User: Convert the point $(0,3)$ in rectangular coordinates to polar coordinates.  Enter your answer in the form $(r,\theta),$ where $r > 0$ and $0 \le \theta < 2 \pi.$

Assistant: $<$think$>$
\end{promptboxmath}

\begin{promptboxmath}{SkillSet inference prompt}
Add all numbers in a list together sequentially. Output type: int. Example: Add all numbers in this list: [1, 2, 3, 4]. Answer: 10

Problem: Add all numbers in this list: [65, 3, 64, 22, 38, 85]

Please solve this problem. Put your final answer within \textbackslash boxed\{\}.
\end{promptboxmath}

\subsection{Compositionality}
\label{app:composition}

\paragraph{Text-based composition prompts (Txt-C and Txt-CoT).}
We use two text-based composition baselines for RQ2 (\Cref{sec:rq2}): \textbf{Txt-C} (text-concise) augments the composite skill prompt with a concise description of the procedural decomposition; \textbf{Txt-CoT} (text-CoT) augments it with a longer chain-of-thought-style decomposition.
Worked examples for both are shown below.

\begin{steerbox}{Txt-C (text-concise)}
Simplify an algebraic expression by combining like terms (format: coefficient*x + constant). Output type: string. Steps: 1) Identify like terms (same variable and power), 2) Add coefficients of like terms, 3) Combine constants. Your final answer must be a simplified expression (e.g., ``21x+13'' or ``5x-3'').

Question: Simplify 1x+4x.

Answer: 5x
\end{steerbox}

\begin{steerbox}{Txt-CoT (text-CoT)}
Simplify an algebraic expression by combining like terms (format: coefficient*x + constant). Output type: string. Steps: 1) Identify like terms (same variable and power), 2) Add coefficients of like terms, 3) Combine constants. Your final answer must be a simplified expression (e.g., ``21x+13'' or ``5x-3'').

Question: Simplify 1x+4x.

Answer: 1. Using Skill \texttt{<compare\_values>}, identify like terms: x-terms=[`1x', `4x'], constants=[]. 2. Using Skill \texttt{<add\_multiple\_numbers>}, combine x-terms: 1+4=5. 3. So the answer is 5x.
\end{steerbox}

\subsection{Personalization Experimental Details}
\label{app:personalization}

\paragraph{Dataset.}
\textsc{SkillSet-Code} comprises 18 LeetCode-style coding problems, each paired with two situational descriptions that favor different algorithmic approaches.
Each problem contributes 4 evaluation records, giving 72 records in total.
Training data for vector extraction consists of 200 model-generated solutions per approach.

\paragraph{Steering Vector Extraction.}
We compute a mean-difference vector for each approach pair $(\mathcal{A}^+,
\mathcal{A}^-)$ as:
\begin{equation}
    v_\ell = \frac{1}{|\mathcal{A}^+|}\sum_{x \in \mathcal{A}^+}
        h_\ell(x)
    \;-\;
    \frac{1}{|\mathcal{A}^-|}\sum_{x \in \mathcal{A}^-}
        h_\ell(x),
\end{equation}
where $h_\ell(x)$ denotes the residual-stream hidden state at layer
$\ell$ for input $x$.

\paragraph{Inference-Time Steering.}
At inference, the steering direction is injected on the last 50\% of layers at the last prefill token and at every generated token:
\begin{equation}
    h_\ell \;\leftarrow\; h_\ell + \alpha\,\hat{v}_\ell,
\end{equation}
where $\hat{v}_\ell$ is $\ell_2$-normalised.
The activation strength reported in~\Cref{sec:rq3} is $\alpha = 1.0$ (see~\Cref{app:extraction-hparams} for the full hyperparameter table and $\alpha$ disclosure).
Generation uses greedy decoding (temperature $= 0$) with a maximum of 4096 tokens.

\paragraph{Evaluation.}
All generated solutions are scored by GPT-4o-mini acting as an LLM judge.
The judge is prompted with the problem statement, the target approach name, and the generated code; it returns a binary verdict on whether the implementation follows the intended approach.
We report \emph{approach match rate}---the fraction of records where the generated code matches the prescribed approach---separately for the baseline (no steering) and steered conditions.
As emphasized in~\Cref{sec:rq3}, this metric measures procedural \emph{controllability}, not functional correctness.

\begin{promptboxmath}{Prompt of LLM-as-a-Judge for Personalization}
\textbf{System:} You are an expert code reviewer. Your task is to determine which algorithmic approach a Python solution uses. Analyze the code structure and logic, not just variable names or comments. Output valid JSON only.

\medskip
\textbf{User:} Given the following coding problem and a generated solution, determine which of the two approaches the solution uses.

\texttt{\#\# Problem}\\
\textit{\{problem\}}

\texttt{\#\# Approach A:} \textit{\{approach\_a\}}\\
\texttt{\#\# Approach B:} \textit{\{approach\_b\}}

\texttt{\#\# Generated Solution}\\
\texttt{```python}\\
\textit{\{code\}}\\
\texttt{```}

Analyze the code's actual algorithmic structure (data structures used, traversal order, control flow) and determine which approach it follows.

Output JSON:
\texttt{```json}\\
\texttt{\{}\\
\quad\texttt{"approach": "A" or "B" or "neither",}\\
\quad\texttt{"confidence": 0.0--1.0,}\\
\quad\texttt{"reasoning": "one sentence explaining why"}\\
\texttt{\}}\\
\texttt{```}
\end{promptboxmath}

\subsection{Evolution Details}
\label{app:evolution}
\paragraph{Evolution Evaluation Prompt}
Below is the LLM-as-a-Judge prompt to evaluate the approach match rate. 
\begin{promptboxmath}{Prompt of LLM-as-a-Judge for Evolution}
\textbf{System:} You are a precise code evaluation judge. Your task is to
read a model output and determine which programming approach it uses or
describes.

\smallskip
\textit{Rules:}
\begin{itemize}
  \item Answer with exactly one word: \texttt{positive}, \texttt{negative},
        or \texttt{none}.
  \item \texttt{positive} means the output uses or clearly describes the
        \textbf{first} (target) approach.
  \item \texttt{negative} means the output uses or clearly describes the
        \textbf{second} approach.
  \item If the output is ambiguous, uses neither approach, or contains only
        a problem description without a clear algorithmic choice, answer
        \texttt{none}.
  \item Base your judgment on actual code, algorithmic choice, or explicit
        approach description in the text.
\end{itemize}

\bigskip
\textbf{User:}

\texttt{Pair:} \{\textit{pair\_key}\} \\
\texttt{POSITIVE approach:} \{\textit{positive\_description}\} \\
\texttt{NEGATIVE approach:} \{\textit{negative\_description}\}

\smallskip
Model output to judge:
\begin{quote}
\{\textit{model\_output}\} \quad \textit{(truncated to 1{,}500 characters)}
\end{quote}

Does this output use the \textsc{positive} or \textsc{negative} approach?\\
\textbf{Answer:} \texttt{positive} / \texttt{negative} / \texttt{none}
\end{promptboxmath}

\begin{figure*}
    \centering
    \includegraphics[width=\linewidth]{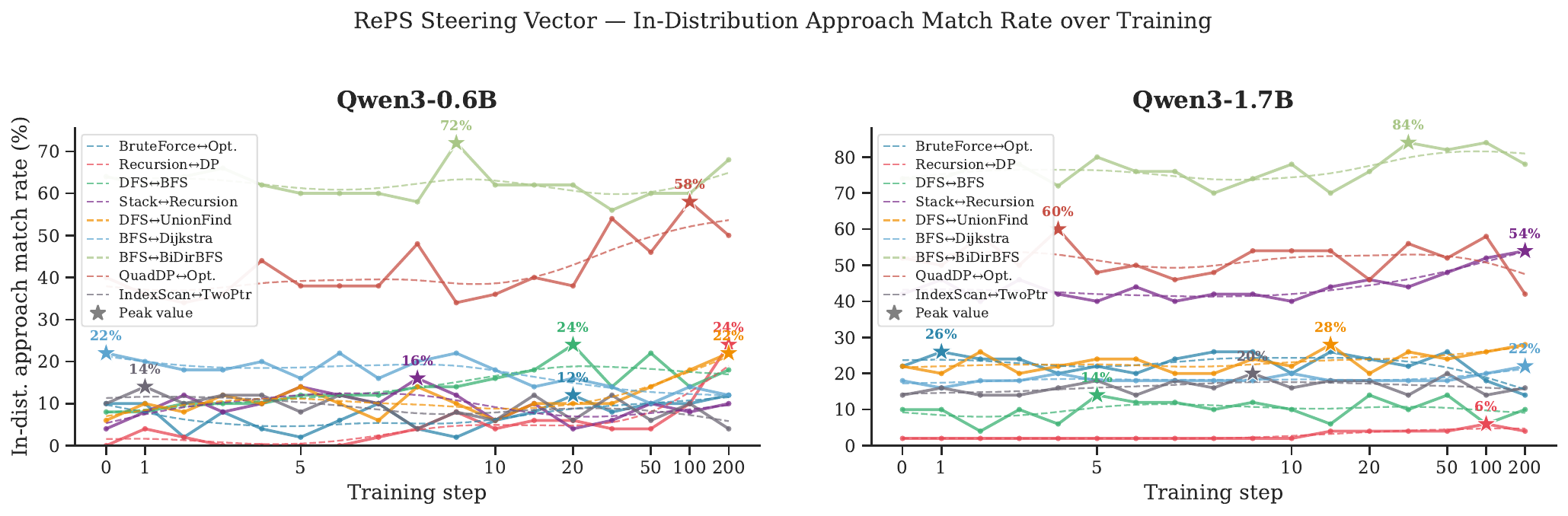}
    \caption{In-Distribution Approach Match Rate Over Training}
    \label{fig:evolution_trajectory}
\end{figure*}

\paragraph{Approach Match Rate Over Training}
Figure~\ref{fig:evolution_trajectory} shows the in-distribution approach match rate over training steps for RePS-optimized steering directions on Qwen3-0.6B (left) and Qwen3-1.7B (right).
Each curve corresponds to an approach pair, with markers indicating peak values.
Performance is consistently non-monotonic, with many pairs reaching their maximum at early or intermediate steps before declining, indicating that the optimal direction emerges before convergence.

\section{Skill-Direction Extraction Procedures}
\label{sec:method-extraction}

We use two complementary extraction schemes, each chosen to match the structure of the corresponding sub-study.

\paragraph{Procedural-execution axis (motivating the single-class extraction).}
We model skill-conditioned activations as concentrating near a low-dimensional manifold along a procedural-execution axis.
Concretely, for a fixed skill $s$ and a layer $\ell$, we assume the hidden state on a skill-$s$ example $x_i$ admits the decomposition
\begin{equation}
\label{eq:procedural-axis}
h_\ell^{(s)}(x_i) \;=\; \mu_\ell^{(s)} \;+\; a_i\, u_\ell^{(s)} \;+\; \epsilon_i,
\end{equation}
where $\mu_\ell^{(s)}$ is the skill-conditioned mean, $u_\ell^{(s)} \in \mathbb{R}^{d_\ell}$ is the procedural-execution direction (the axis along which skill-$s$ execution varies), $a_i \in \mathbb{R}$ is a per-example scalar reflecting how strongly that procedure is engaged on $x_i$, and $\epsilon_i$ is an instance-specific residual.
Under this decomposition, the leading principal component of the centered skill-conditioned activations $\{h_\ell^{(s)}(x_i) - \mu_\ell^{(s)}\}_i$ recovers the procedural-execution direction $u_\ell^{(s)}$ up to a sign.
This view is what motivates the single-class PCA extraction below: in many of our math skills (e.g., \texttt{add\_two\_single\_digits}) there is no natural negative class to form a contrastive direction, so the procedural-execution axis is the most direct target for single-class extraction.

\paragraph{Single-class PCA (RQ1/2, \textsc{SkillSet-Math}).}
For each math skill $s$ (atomic or composite), we collect up to 500 training examples per skill and record the hidden state at every layer, taken at the \emph{last token} position of the formatted input.
Each formatted input consists of the skill instruction, the problem, and the corresponding answer text.
For atomic skills the answer text is the final numeric answer; for composite skills the answer text is the full step-by-step derivation (i.e., a longer answer that exposes the procedural composition).
We center these per-skill activations and compute the leading principal component (top-1 PCA), giving one $\ell_2$-normalized direction $v_\ell^{(s)}$ per layer.
This single-class, positive-only extraction targets the procedural-execution direction $u_\ell^{(s)}$ in~\Cref{eq:procedural-axis}: for many math skills there is no natural negative class with which to construct a contrastive direction (e.g., the negation of \texttt{add\_two\_single\_digits} is not a single, well-defined alternative procedure), so the procedural-execution axis is the most direct target.
We disclose two methodological details that read this design more carefully.
First, because the prompt used for extraction includes the correct answer, the extracted direction reflects the hidden state of the model conditioned on observing problem-and-answer together, not the hidden state of the model when it is generating the answer on its own.
Second, the principal-component sign is indeterminate; we use the raw PCA direction without sign correction, and the sign of $\alpha$ in evaluation is fixed across skills.

\paragraph{Contrastive mean-difference (RQ3, \textsc{SkillSet-Code}).}
For each algorithmic-strategy pair (e.g., DFS vs.\ BFS), we form 200 positive (target-strategy) and 200 negative (alternative-strategy) solutions per pair.
We record the hidden state at the last token of each formatted input (\verb|Problem: ...\n\nSolution:\n<code>|) at every layer, and compute the mean-difference direction
$
v_\ell = \mu_\ell^{+} - \mu_\ell^{-},
$
which we then $\ell_2$-normalize.
Because the pair is itself contrastive, this construction yields a sign-determinate direction (pos $\to$ neg) and a natural negative class that the single-class PCA extraction lacks.

\paragraph{Shared inference-time choices.}
Across both extractions, we use \emph{per-layer} directions, intervene on the last 50\% of layers (e.g., layers 14--27 for Qwen3-0.6B's 28-layer stack), and add the direction to hidden states at the last prefill token and at every generated token via $\tilde h_\ell \leftarrow h_\ell + \alpha v_\ell$.
We report the activation strength $\alpha$ used in each sub-study explicitly:
RQ1/2 use $\alpha = 2.0$;
RQ3 evaluations in this paper report results at $\alpha = 1.0$, which is the value used in the tracker runs underlying the reported numbers (the configured sweep grid was $\{0, 0.5, 1, 2, 4\}$).
Full hyperparameter details (token positions, sample counts, prompt formats, target-layer scope, $\alpha$ grids) are provided in~\Cref{app:extraction-hparams} below.

\paragraph{Why two schemes.}
The use of two extraction schemes is principled rather than ad hoc: single-class PCA matches the structure of atomic/composite math skills (one positive class, no natural negative), while mean-difference matches the structure of paired algorithmic strategies (two structured classes whose contrast \emph{is} the skill of interest).
This pairing of extraction scheme to the structural form of the skill is essential to our claims; in particular, in RQ1/2 we do not claim PCA is the only or best extractor, only that it provides a single, consistent procedure with which to study the four observable properties on \textsc{SkillSet-Math}.

\section{Skill-Direction Extraction Hyperparameters}
\label{app:extraction-hparams}

This section provides the full hyperparameter specification for the two extraction schemes described in~\Cref{sec:method-extraction}.
We describe these in methodological terms (token positions, sample counts, prompt formats, target-layer scope, $\alpha$ values).

\paragraph{RQ1/RQ2 setting (single-class PCA on \textsc{SkillSet-Math}).}
\Cref{tab:hparams-pca} summarizes the extraction settings used for steerability and composition.

\begin{table*}[htbp]
\centering
\small
\setlength{\tabcolsep}{6pt}
\renewcommand{\arraystretch}{1.15}
\begin{tabular}{ll}
\toprule
\textbf{Item} & \textbf{Setting} \\
\midrule
Extraction method        & Single-class PCA (positive samples only; non-contrastive) \\
Principal component      & First principal component (top-1) \\
Layer scope              & Per-layer (independently extracted at every layer) \\
Token position           & Last token of the formatted input \\
Sample size              & Up to 500 examples per skill (capped at available count) \\
Prompt format            & Instruction + Problem + Answer \\
Answer text (atomic)  & Final answer \\
Answer text (composite)& Full step-by-step derivation \\
Normalization            & $\ell_2$ normalization of the direction \\
Skill inventory          & \textsc{SkillSet-Math} (16 atomic + 18 composite = 34 skills) \\
Evaluation datasets      & \textsc{SkillSet-Math}; transfer to GSM8K and MATH500 \\
Intervention             & Additive ($\tilde h_\ell = h_\ell + \alpha v_\ell$) \\
Activation strength      & $\alpha = 2.0$ \\
Target layers            & Last 50\% of layers (e.g., layers 14--27 on Qwen3-0.6B) \\
Trigger tokens           & Last prefill token + every generated token \\
\bottomrule
\end{tabular}
\caption{RQ1/RQ2 extraction setting: single-class PCA on \textsc{SkillSet-Math}.}
\label{tab:hparams-pca}
\end{table*}

\paragraph{RQ3 setting (contrastive mean-difference on \textsc{SkillSet-Code}).}
\Cref{tab:hparams-meandiff} summarizes the extraction settings used for personalization.

\begin{table*}[htbp]
\centering
\small
\setlength{\tabcolsep}{6pt}
\renewcommand{\arraystretch}{1.15}
\begin{tabular}{ll}
\toprule
\textbf{Item} & \textbf{Setting} \\
\midrule
Extraction method        & Mean-difference contrastive: $v = \mu^{+} - \mu^{-}$ \\
Layer scope              & Per-layer (independently extracted at every layer) \\
Token position           & Last token of the formatted input \\
Normalization            & $\ell_2$ normalization of the direction \\
Sample size              & 200 positive + 200 negative per approach pair \\
Prompt format            & ``Problem: \ldots\textbackslash n\textbackslash nSolution (Approach):\textbackslash n\textit{<code>}'' \\
Skill inventory          & 9 algorithmic-strategy pairs (e.g., DFS vs.\ BFS) \\
Evaluation set           & 72 evaluation records \\
Intervention             & Additive ($\tilde h_\ell = h_\ell + \alpha v_\ell$) \\
Target layers            & Last 50\% of layers \\
Trigger tokens           & Last prefill token + every generated token \\
\bottomrule
\end{tabular}
\caption{RQ3 extraction setting: contrastive mean-difference on \textsc{SkillSet-Code}.}
\label{tab:hparams-meandiff}
\end{table*}

\paragraph{Shared inference-time choices.}
Both extractions use per-layer directions, intervene on the last 50\% of layers, and trigger at the last prefill token and at every generated token.
Generation uses greedy decoding (temperature $= 0$) with a maximum of 4096 tokens.

% \paragraph{Note on $\alpha$ for RQ3.}
% The configured sweep grid for RQ3 is $\{0, 0.5, 1.0, 2.0, 4.0\}$.
% The numbers reported in~\Cref{sec:rq3} ( \Cref{tab:skillchoice24_category}) correspond to $\alpha = 1.0$, the value actually used in the tracker runs from which those numbers were drawn.
% Earlier draft prose referenced $\alpha = 2.0$; we follow the value actually used at the time of the reported run, and disclose this here for reproducibility.

\paragraph{Disclosure: RQ1/2 prompts include the correct answer.}
For RQ1 and RQ2, the formatted input from which PCA directions are extracted includes the instruction, the problem, and the correct answer (final answer for atomic skill; full step-by-step derivation for composite skill).
The extracted direction therefore reflects the model's hidden state conditioned on observing problem-and-answer together, not its hidden state when generating the answer autonomously.
We make this design choice explicit because it materially affects how the extracted direction should be interpreted: the direction targets the procedural-execution axis present when the model is conditioned on a worked example, and is then re-injected on unseen problems at inference time.

\section{Cross-Family Results on Llama-2-7B-chat}
\label{app:llama-crossfamily}

All experiments in the main paper use the Qwen3 family, which shares a single architecture and training recipe.
To test whether the reported properties depend on that choice, we repeat part of the controlled study on Llama-2-7B-chat, a 2023-era model with a different tokenizer, training corpus, and instruction-tuning procedure.
We cover skill composition on all 18 composite skills (\Cref{tab:llama-composition}), transfer to GSM8K and MATH500 (\Cref{tab:llama-transfer}), and a sweep over intervention strength on five representative skills (\Cref{tab:llama-alpha}).
Our goal is not a new generic steering method, but a controlled framework for latent procedural skills; these results extend the evidence beyond Qwen3.

\paragraph{Skill Composition.}
The central composition result of \Cref{sec:rq2} reproduces across model families.
Vector composition outperforms both text-based composition baselines on 13 of 18 composite skills, the same ratio observed on Qwen3-0.6B, and attains the highest average accuracy (31.5, versus 21.6 for Txt-C and 20.7 for Txt-CoT).
Because Llama-2-7B-chat differs from Qwen3 in tokenizer, training corpus, and instruction-tuning procedure, this provides direct evidence that the composition finding is not limited to the Qwen3 model family.
Absolute accuracies are substantially lower than on Qwen3, consistent with the weaker base model.

\begin{table*}[htbp]
\centering
\small
\setlength{\tabcolsep}{4pt}
\begin{tabular}{lrrrr}
\toprule
\textbf{Composite skill} & \textbf{Base} & \textbf{Txt-C} & \textbf{Txt-CoT} & \textbf{Vec} \\
\midrule
\texttt{multi\_digit\_addition}        & 84 &  7 & 24 & \textbf{85} \\
\texttt{multi\_digit\_subtraction}     & \textbf{74} & 11 & 21 & 71 \\
\texttt{multi\_digit\_multiplication}  & \textbf{11} &  1 &  1 & \textbf{11} \\
\texttt{long\_division}                & 58 & 19 & 18 & \textbf{65} \\
\texttt{decimal\_addition}             & \textbf{78} & 65 & 36 & 77 \\
\texttt{sum\_of\_range}                &  5 &  1 &  2 & \textbf{7} \\
\texttt{rounding\_numbers}             & 29 & 37 & \textbf{40} & 31 \\
\texttt{percentage\_calculation}       & 29 & 28 & 18 & \textbf{32} \\
\texttt{greatest\_common\_divisor}     & 13 & \textbf{23} & 16 & 11 \\
\texttt{solve\_linear\_equation}       & 90 & 54 & 49 & \textbf{93} \\
\texttt{solve\_system\_two\_equations} &  0 &  0 &  0 & \textbf{1} \\
\texttt{check\_equation\_validity}     &  0 & \textbf{55} & 40 &  0 \\
\texttt{inequality\_solving}           & \textbf{1} &  0 &  0 & \textbf{1} \\
\texttt{algebraic\_simplification}     & 25 &  8 & 19 & \textbf{26} \\
\texttt{range\_validation}             & 14 & 39 & \textbf{43} & 10 \\
\texttt{linear\_search}                &  6 & \textbf{29} & 21 &  5 \\
\texttt{reverse\_list}                 & \textbf{11} &  2 &  2 & 10 \\
\texttt{count\_occurrences}            & 27 &  9 & 23 & \textbf{31} \\
\midrule
\textbf{Average}                       & 30.8 & 21.6 & 20.7 & \textbf{31.5} \\
\bottomrule
\end{tabular}
\caption{Skill composition on Llama-2-7B-chat. Accuracy (\%) on the 18 composite skills of \textsc{SkillSet-Math}. Bold marks the best method per skill.}
\label{tab:llama-composition}
\end{table*}

\begin{table*}[t]
\centering
\small
\setlength{\tabcolsep}{4pt}
\begin{tabular}{lrrrrr}
\toprule
\textbf{Skill} & \textbf{Base} & $\alpha{=}0.5$ & $\alpha{=}1.0$ & $\alpha{=}2.0$ & \textbf{Best $\Delta$} \\
\midrule
\texttt{subtract\_two\_single\_digits} & 84 & 88 & 88 & \textbf{95} & $+11$ \\
\texttt{multi\_digit\_subtraction}     & 74 & 78 & 79 & \textbf{84} & $+10$ \\
\texttt{inequality\_solving}           &  1 &  0 &  1 & \textbf{10} & $+9$ \\
\texttt{compare\_values}               & 51 & 53 & 53 & \textbf{60} & $+9$ \\
\texttt{solve\_linear\_equation}       & 90 & 91 & 92 & \textbf{95} & $+5$ \\
\bottomrule
\end{tabular}
\caption{Sensitivity to intervention strength $\alpha$ on Llama-2-7B-chat, for five representative skills under single-class PCA extraction. Accuracy (\%).}
\label{tab:llama-alpha}
\end{table*}

\paragraph{Transfer to Real Benchmarks.}
Skill directions extracted on \textsc{SkillSet-Math} also transfer positively on Llama-2-7B-chat, improving GSM8K by $3.26$ and MATH500 by $2.80$ points over zero-shot decoding.
These gains are considerably larger than those observed on the stronger Qwen3 models in \Cref{tab:math_datasets_steering}, which we attribute to a headroom effect: an inference-time procedural intervention can only correct errors the base model actually makes, and Llama-2-7B-chat starts from a far weaker baseline ($25.25$ on GSM8K, $6.00$ on MATH500) than Qwen3-14B ($95.83$ and $92.40$).
This pattern is consistent with recent scaling work on activation steering, which reports that steering effects can diminish with model size~\citep{ali2025scaling}.

\begin{table}[htbp]
\centering
\small
\begin{tabular}{lrrr}
\toprule
\textbf{Benchmark} & \textbf{0-shot} & \textbf{Steered} & \textbf{$\Delta$} \\
\midrule
GSM8K   & 25.25 & 28.51 & $+3.26$ \\
MATH500 &  6.00 &  8.80 & $+2.80$ \\
\bottomrule
\end{tabular}
\caption{Transfer of \textsc{SkillSet-Math} skill directions on Llama-2-7B-chat. Accuracy (\%).}
\label{tab:llama-transfer}
\end{table}

\paragraph{Sensitivity to Intervention Strength.}
The activation strength $\alpha$ is an intervention hyperparameter rather than a universal constant, and prior steering work likewise tunes it per setting rather than fixing one global coefficient~\citep{panickssery2023steering,DBLP:Tonebank}.
This is expected in our setting because the sub-studies use different benchmarks and different vector definitions: single-class PCA on \textsc{SkillSet-Math} versus contrastive mean-difference on \textsc{SkillSet-Code}.
We therefore report a consistent $\alpha$ within each dataset and extraction regime rather than forcing a single value across all sub-studies.
\Cref{tab:llama-alpha} shows the corresponding sweep on Llama-2-7B-chat: the useful range is skill-dependent, with \texttt{inequality\_solving} responding only at $\alpha{=}2.0$ while \texttt{solve\_linear\_equation} varies little across the sweep, and large $\alpha$ can over-steer the model.
These results support calibrating intervention strength to the model, representation scale, and target behavior.

\end{document}